\documentclass[10pt]{article} 
\usepackage[preprint]{tmlr}

\usepackage{amsmath,amsfonts,bm}

\def\eqref#1{equation~\ref{#1}}

\def\1{\bm{1}}

\DeclareMathAlphabet{\mathsfit}{\encodingdefault}{\sfdefault}{m}{sl}
\SetMathAlphabet{\mathsfit}{bold}{\encodingdefault}{\sfdefault}{bx}{n}

\usepackage{hyperref}
\usepackage{url}
\usepackage[utf8]{inputenc} 
\usepackage[T1]{fontenc}    
\usepackage{booktabs}       
\usepackage{amsfonts}       
\usepackage{nicefrac}       
\usepackage{microtype}      
\usepackage{xcolor}         
\usepackage{graphicx}
\usepackage{subfig}
\usepackage{natbib}
\usepackage{amsmath}
\usepackage{algorithm}
\usepackage{algpseudocode}
\usepackage{cancel}
\usepackage{amssymb}
\usepackage{mathtools}
\usepackage{amsthm}
\usepackage{caption}
\usepackage{subcaption}

\title{Model Alignment Search}

\author{\name Satchel Grant \email grantsrb@stanford.edu \\
    \addr Departments of Psychology and Computer Science \\
    Stanford University
}

\def\month{07}  
\def\year{2025} 
\def\openreview{\url{https://openreview.net/forum?id=YPnYpiru5W}} 

\begin{document}

\newcommand{\dmodel}{128}
\newcommand{\dmodelhalf}{64}
\newcommand{\nepochs}{500}
\newcommand{\batchsize}{128}
\newcommand{\maxcount}{20}
\newcommand{\holdouts}{4, 9, 14, and 17}
\newcommand{\hidsize}{512}
\newcommand{\dropout}{0.5}
\newcommand{\nonlinearity}{GELU}
\newcommand{\learnrate}{0.001}
\newcommand{\ntrain}{1000}
\newcommand{\nval}{500}
\newcommand{\nseeds}{2}
\newcommand{\seqlen}{43} 

\newcommand{\fullxmas}{Counterfactual Latent MAS}
\newcommand{\xmas}{CLMAS}
\newcommand{\stitch}{Stitch}

\newcommand{\ndemotypes}{3}
\newcommand{\singleobject}{Single-Object}
\newcommand{\sameobject}{Same-Object}
\newcommand{\multiobject}{Multi-Object}
\newcommand{\varylen}{Variable-Length}
\newcommand{\varyprob}{0.2}
\newcommand{\arithmetic}{Arithmetic}
\newcommand{\modulo}{Modulo}
\newcommand{\modulus}{4}
\newcommand{\rounding}{Rounding}
\newcommand{\roundmultiple}{3}
\newcommand{\arithmaxnew}{9}
\newcommand{\arithnumbase}{10}
\newcommand{\aritmaxcumu}{100}

\newcommand{\countupdown}{Up-Down}
\newcommand{\countupup}{Up-Up}
\newcommand{\varcount}{Count}
\newcommand{\varcumuval}{Cumu Val}
\newcommand{\varremops}{Rem Ops}
\newcommand{\varphase}{Phase}
\newcommand{\vardemocount}{Demo Count}
\newcommand{\varrespcount}{Resp Count}
\newcommand{\varinptval}{Last Value}
\newcommand{\varfull}{Full}
\newcommand{\distrsoln}{MapReduce}
\newcommand{\inptval}{Tok Val}
\newcommand{\tokhist}{Tok History}
\newcommand{\eosreadout}{EOS Readout}

\newcommand{\dasntrain}{10000}
\newcommand{\dasnval}{1000}
\newcommand{\dasbatchsize}{512}
\newcommand{\daslr}{0.001}
\newcommand{\dasnepochs}{1000}
\newcommand{\dasdims}{32}
\newcommand{\lowdasdims}{16}
\newcommand{\highdasdims}{32}


\maketitle

\begin{abstract}
When can we say that two neural systems perform a task in the same way? What nuances do we miss when we fail to causally probe the representations of the systems, and how do we establish bidirectional causal relationships? In this work, we introduce a method that bidirectionally transfers neural activity between artificial neural networks and uses their resulting behavior as a measure of functional similarity. We first show that the method can be used to transfer the behavior from one frozen Neural Network (NN) to another in a manner similar to model stitching, and we show how the method can differ from correlative similarity measures like Representational Similarity Analysis. Next, we empirically and theoretically show how the method can be equivalent to model stitching when desired, or it can take a form that has a more restrictive focus to shared causal information; in both forms, it reduces the number of required matrices for a comparison of n models to be linear in n. We then present a case study on number-related tasks showing that the method can be used to examine specific subtypes of causal information demonstrating that numbers can be encoded differently in recurrent models depending on the task, and we present another case study showing that MAS can reveal misalignment in fine-tuned DeepSeek-r1-Qwen-1.5B models. Lastly, we augment the loss function with a counterfactual latent (CL) auxiliary objective to improve causal relevance when one of the two networks is causally inaccessible (as is often the case in comparisons with biological networks). We use our results to encourage the use of causal methods in neural similarity analyses and to suggest future explorations of network similarity methodology for model misalignment.
\end{abstract}

\section{Introduction}

An important question for understanding both Artificial and Biological Neural
Networks (ANNs and BNNs) is knowing what it means for one distributed system
to model or represent another \citep{sucholutsky2023repalign}. 
Establishing isomorphisms between different distributed systems can
be useful for simplifying their complexity and for understanding
otherwise opaque inner mechanisms. Instead of asking how to interpret neural
activity directly, we can instead compare the internals of the simplified neural systems, or complex neural systems
understood through a higher level of analysis (i.e. architecture,
learning rule, training data), to the internals of the unknown systems
\citep{cao_explanatory_2021,cao_explanatory_2024,richards2019dlframework}.
In addition, there are several open questions about how
representations differ or converge between architectures, tasks,
and modalities
\citep{huh2024platonic,sucholutsky2023repalign,wang2024universality,li_vision_2024,Hosseini2024universality,zhang2024assessinglearningalignmentunimodal,grant2024das}.
\begin{figure*}[t!]
    \centering
    \includegraphics[width=0.8\textwidth]{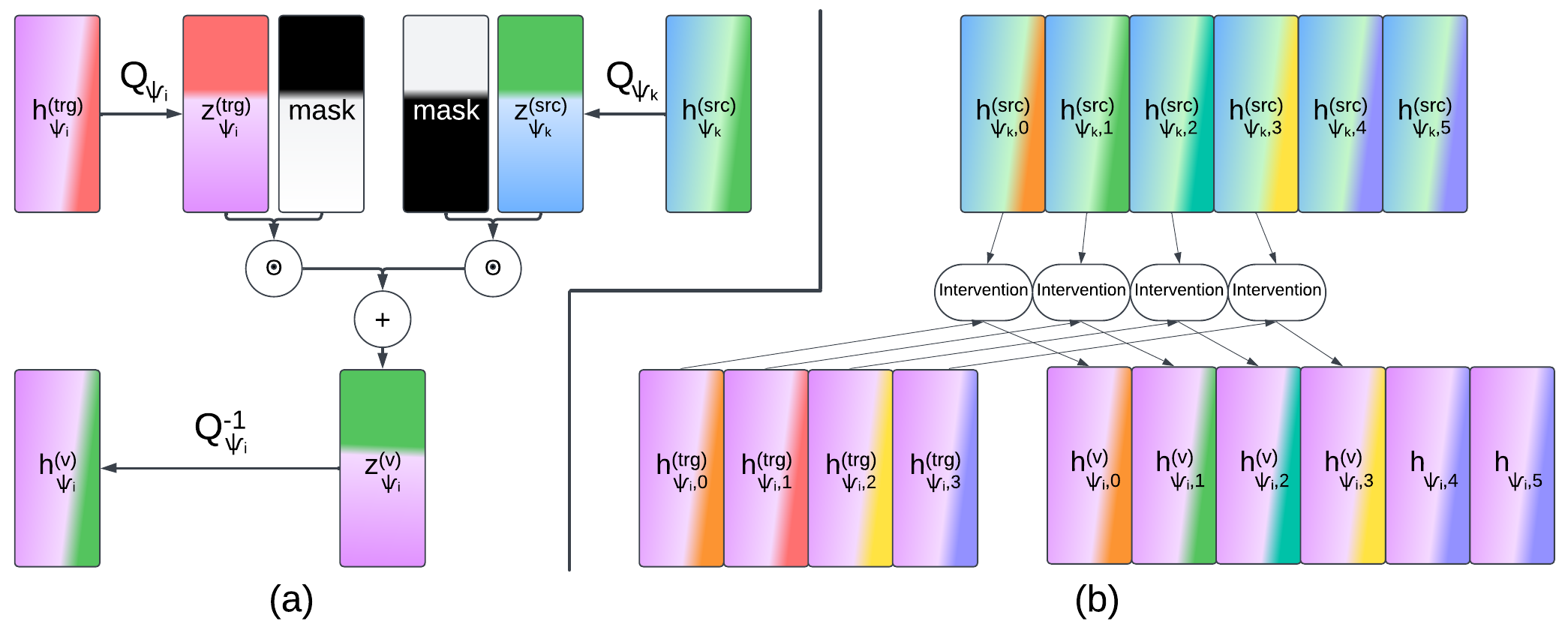}
    \caption{
    \textbf{(a)} A depiction of an interchange intervention from Equation~\ref{eq:masinterchange} on the target latent
    vector $h^{(trg)}_{\psi_i}$ from model $\psi_i$ using the
    source latent vector $h^{(src)}_{\psi_k}$ from $\psi_k$.
    Rectangles represent vectors; colors distinguish between behaviorally relevant and extraneous activity. The causally relevant
    information is spread across the neural population \citep{Smolensky1988,park2023linearrephypoth}, represented by
    the red and green semi-vertical slices in the respective $h$
    vectors. To disentangle and isolate the behavioral null-space,
    we rotate the $h$ vectors into an aligned space, using $Q_{\psi_i}$ and
    $Q_{\psi_k}$, where the behaviorally relevant information is organized along separate dimensions than
    the extraneous, behavioral null-space. We can then intervene on and transfer the
    relevant information without affecting other information. In the figure, this is done by
    applying binary masks (black represents 0s and white 1s and
    $\odot$ is a Hadamard product) to the $z$ vectors and taking
    their sum. We then invert the rotation on $z^{(v)}_{\psi_i}$ to
    return it to the original latent space where it can be used by $\psi_i$ to make predictions.
    \textbf{(b)} Depicts Stepwise MAS where the individual intervention shown in
    panel (a) is applied at multiple token positions in the sequence.
    We limit our interventions to contiguous sets of tokens starting from the first
    token and ending with a sampled position $t$.
    }
    \label{fig:masintervention}
\end{figure*}

There are many existing methods for analyzing the similarity between
different neural representations. Among these, there exist both correlational and causal methods. We see
examples of the correlational variety in works that perform direct
correlational analyses between
individual ANN activations and BNN firing rates
\citep{Yamins2016,Maheswaranathan2019,khosla2023softmatching,williams2022procrustes},
and in works that use Representational Similarity Analysis (RSA) \citep{kriegeskorte2008rsa}---or Centered Kernel
Alignment (CKA) \citep{kornblith2019cka,williams2024rsacka}---finding 2nd
order isomorphisms between model and system.
We also see examples of this in linear decoding techniques, where linear
decodability and predictability can be used as a measure of the type of
information encoded in distributed representations
\citep{chen2020simclr,radford2021clip,grill2020byol,caron2021dino,Haxby2001ogmvpa,Haxby2013mvpareview,lampinen2024learned,thobani2024interanimal}.
However, such correlative methods have received many criticisms for their
sensitivity to noise structure (e.g. nuisance variables), reliance on
outliers and large principal components, sensitivity to transformations
that preserve functional behavior, and sensitivity to sample sizes
and vector dimensionality \citep{Dujmovic2022rsaprobs,popal2020rsaguide,davari2022ckareliability,murphy2024biasedcka}.

What do we miss when we ignore the functional, behavioral relevance of
the neural activity?
Many works have pointed out the disconnect between neural activity
and its causal impact on network behavior
\citep{pearl2010causalmediation,geiger2024causalabstractiontheoreticalfoundation,
cloos2024differentiableoptimizationsimilarityscores,schaeffer2024doesregressbrain,lampinen2025representation}.
How do we incorporate a focus on behavioral relevance in our
measures of neural similarity?
Some works have emphasized network function in representational
comparisons by transforming intermediate representations from one neural
system into a usable form for another. One such method for
performing such operations is known as \emph{model stitching}.
We see examples of this in
ANN-BNN comparisons in works like \cite{sexton_reassessing_2022}
where they use transformed neural recordings in a trained
vision model, and in ANN-ANN comparisons where stitching has been
used both as a measure of functional representational similarity and
for improving one of the two models
\citep{lahner2023directalignment,moschella_relative_2023,bansal2021modelstitching,lenc2015modelstitching,klabunde2025similaritysurvey,braun2025functionaldissociation}.
However, model stitching has its own limitations.
\citet{hernandez2022stitchinglayerdiffs} showed that later model
layers can successfully transfer behavior regardless of distance from
the receiving layer, and \citet{smith2025stitchingfailures} showed
that stitched representations can have lower intrinsic
dimensionality than native representations, and networks trained on
notably different tasks can be successfully stitched together.

These shortcomings leave us with the broad question: how do we
determine when neural systems are the same? In this work, we will
pursue a notion of representational similarity that establishes
an isomorphism between representational spaces with emphasis
on preserving behavior. This means that we desire a mapping
between model representations that is bidirectional (invertible),
and we want the success of the mapping to be measured, at least
in part, with respect to the networks' function (i.e. behavior).
In this work, we introduce such a method that uses causal
interventions to isolate and compare behaviorally relevant activity
from neural representations in different ANNs. The method has a
close relationship to Distributed Alignment Search (DAS) \citep{geiger2021,geiger2023boundless,wu2024alpacadas} for which
we name it \emph{Model Alignment Search} (MAS).
MAS learns a rotation matrix, or more generally an alignment function,
for each model in the comparison to simultaneously uncover causally
relevant latent subspaces in each model and map these spaces onto one
another so that behavioral information can be patched between models.
The networks' resulting behavior following an intervention can then be
compared to the expected counterfactual behavior as a means of
measuring functional similarity.

The contributions of this work are as follows.
\begin{enumerate}
    \item We introduce and motivate MAS by comparing it to correlative
    representational similarity methods demonstrating a need
    to supplement correlational with causal methods depending on the
    research goals.
    \item We empirically and theoretically compare MAS to model stitching,
    showing that it requires less compute when comparing 3 or more models,
    and it can be more restrictive to causal subspaces than model
    stitching.
    \item Using numeric cognition as a case study, we show how MAS can be used
    to examine the similarity of specific types of behavioral information in
    neural representations.
    \item Using model misalignment as a case study, we show that the method can be used to reveal toxicity in
    DeepSeek-R1-Qwen-1.5B models \citep{guo2025deepseek} finetuned to be toxic or nontoxic.
    \item Lastly, we introduce a \emph{counterfactual latent} auxiliary loss objective
    that can be used to find behaviorally relevant alignments in cases where
    one of the two models is causally inaccessible (making
    the technique potentially relevant for comparisons between ANNs and BNNs).
\end{enumerate}
\section{Background and Related Work}

There has been a call to use causal methods for both
interpreting neural activity and making comparisons between
different NN internals \citep{geiger2024causalabstractiontheoreticalfoundation, sexton_reassessing_2022,feather2025brain,lampinen2025representation}.
This call has been made due to principles of causal mediation \cite{pearl2010causalmediation}, and, more recently,
representation and causal computation have been shown to be
different in NNs \citep{hermann2020shapes,braun2025functionaldissociation,lampinen2025representation}.
In the mechanistic interpretability literature, a popular
method for making claims
about cause and effect in NNs is activation patching
\citep{geiger2020neural,vig2020causal,wang2022patching,meng2023activpatching} where experimentalists
manipulate the NN activity to observe downstream effects.
One such method is DAS, which finds causal subspaces that align
with high-level variables from causal abstractions
\citep{geiger2021,geiger2023boundless,wu2024alpacadas}.
In cases of multi-model comparisons, a particularly popular method
for causally
comparing representations between different models is model stitching
\citep{lenc2015modelstitching,bansal2021modelstitching}. In this
section, we present a background on both model stitching and DAS to
establish notation and a conceptual framework for MAS.

\subsection{Model Stitching}\label{sec:stitchingformulation}
Model stitching
aims to learn a mapping from the latent state space of one
NN, $\psi_1$, to that of another, $\psi_2$. Let
$h_{\psi_i} \in \mathbb{R}^{d_{\psi_i}}$ denote a latent representation
from $\psi_i; \, i\in\{1,2\}$. Stitching seeks a transformation
$T : \mathbb{R}^{d_{\psi_1}} \to \mathbb{R}^{d_{\psi_2}}$ such that
\[
T\!\left(h_{\psi_1}\right) \approx h_{\psi_2}
\]
where $h_{\psi_2} \in \mathbb{R}^{d_{\psi_2}}$ is the corresponding
representation in $\psi_2$. The mapped representation $T(h_{\psi_1})$
is then fed into $\psi_2$'s remaining layers, denoted $f_{\psi_2}(x,h)$,
so that $\psi_2$ can make
predictions consistent with those produced by $\psi_1$.

Although $T$ can be trained in a number of ways, we focus on cases in
which $T$ is learned by optimizing $\psi_2$'s outputs, conditioned
on the stitched representation $\hat{h}_{\psi_2} = T(h_{\psi_1})$,
to match $\psi_1$'s behavior, also conditioned on $h_{\psi_1}$. The learning
objective is a behavioral matching loss:
\begin{equation} \label{eq:mstitchloss}
\mathcal{L}(T) \;=\;
\mathbb{E}_{x \sim \mathcal{D}} \big[ \ell\big( f_{\psi_1}(x,h_{\psi_1}) \,\|\, f_{\psi_2}(x,\hat{h}_{\psi_2}) \big) \big]
\end{equation}
where $x\sim \mathcal{D}$ are samples from a dataset and $\ell$ is a divergence measure
(e.g., KL divergence or cross-entropy)
between $\psi_1$'s output distribution and $\psi_2$'s output
distribution under the stitched representation.

In this work, we narrow our focus to transformations of the form
\[
T(h_{\psi_1}) = a W h_{\psi_1}
\]
where $a \in \mathbb{R}$ is a scalar constant and
$W \in \mathbb{R}^{d_{\psi_2} \times d_{\psi_1}}$ is a matrix—possibly
low rank—with orthonormal rows.

Model stitching has been used as a method to make claims
about the similarity between two different NNs
\cite{bansal2021modelstitching}. However, it has problems as a
similarity method due to the fact that it is
unidirectional, and, as we will show in Section~\ref{res:stitching},
it can make use of $\psi_1's$ behavioral null
space when mapping to $\psi_2$---where the behavioral null space is
defined as the span of vector directions that do not change the
NN's behavior. Furthermore, a successful mapping can ignore naturally
occurring variability in $\psi_2$'s representations
\citep{smith2025stitchingfailures}. Thus,
a successful mapping $T$ only needs to produce a small set of
sufficient representations for any given behavior meaning that
successful $T$'s do not necessarily indicate that $\psi_1$
and $\psi_2$ perform the task in the same way.

\subsection{DAS Formulation}\label{sec:DASformulation}
DAS is a framework for causally determining the degree of alignment
between a single NN's latent representations and high-level variables from a causal
abstraction \citep{geiger2021,geiger2023boundless,wu2024alpacadas}. We
can think of such variables as those in a computer
program, or nodes in a Directed Acyclic Graph (DAG), that can take
on a well defined set of values. DAS does this
alignment by attempting to transform model latent states $h \in \mathbb{R}^{d_\psi}$
into aligned vectors $z \in \mathbb{R}^{d_\psi}$ that consist of
contiguous subspaces that have an analogous causal effect on behavior
as the high-level, causal variables from a pre-specified causal
abstraction. The
transformation is performed by learnable, invertible \emph{Alignment
Function} (AF), $\mathcal{A}$, as follows: $z = \mathcal{A}(h)$
\citep{grant2024das}.
We narrow our focus to orthogonal AFs: $\mathcal{A}(h) = Qh$ where
$Q\in\mathbb{R}^{d_\psi \times d_\psi}$ is an orthogonal matrix.
The purpose of this transformation is that
it organizes causal subspaces into separated, contiguous dimensions
in $z$. This allows for causal interventions on the
values of specific subspaces---analogously to changing value of a
variable in a computer program---without affecting the values of the
other subspaces.

Concretely, for a given causal abstraction with variables
$\text{var}_i \in \{\text{var}_1, \text{var}_2, ..., \text{var}_n\}$,
DAS learns a rotation $Q$ to find a $z$ composed of subspaces
$\Vec{z}_{\text{var}_i} \in \mathbb{R}^{d_{\text{var}_i}}$ corresponding to each
of the variables of the causal abstraction. We also include a causally
irrelevant subspace, $\Vec{z}_{\text{extra}}\in \mathbb{R}^{d_{\text{extra}}}$,
encoding extraneous, behaviorally irrelevant activity (i.e., the
behavioral null-space and dormant subspaces---defined as those that do not vary
between inputs \citep{makelov2023interpillusions}).
\begin{eqnarray}\label{eq:zdef}
    Q(h) = z = \begin{bmatrix}
        \Vec{z}_{\text{var}_1} \\
        \Vec{z}_{\text{var}_2} \\
        \cdots \\
        \Vec{z}_{\text{var}_n} \\
        \Vec{z}_{\text{extra}}
    \end{bmatrix}
\end{eqnarray}
Each $\Vec{z}_{\text{var}_i} \in \mathbb{R}^{d_{\text{var}_i}}$ is a
column vector of
potentially different lengths satisfying the relation
$d_{\text{extra}} + \sum_{i=1}^n d_{\text{var}_i} = d_\psi$.
Under this assumption, the value of a single causal variable encoded in
$h$ can be freely exchanged by performing an interchange intervention
defined as follows:
\begin{equation}\label{eq:interchange}
    h^{(v)} = Q^{-1}(
        (1-D_{\text{var}_i}) Qh^{(trg)} +
        D_{\text{var}_i} Qh^{(src)}
    )
\end{equation}
Where $D_{\text{var}_i}\in \mathbb{R}^{d_{\psi}\times d_{\psi}}$ is a manually chosen,
diagonal, binary matrix with $d_{\text{var}_i}$ contiguous ones
along the diagonal used to isolate the dimensions that make up
$\Vec{z}_{\text{var}_i}$,
$h^{(src)}$ is the \emph{source vector} from which the subspace
activity is harvested, $h^{(trg)}$ is the \emph{target vector} into
which the harvested activations are substituted, and $h^{(v)}$ is the
resulting intervened vector that is used to replace $h^{(trg)}$ in the
model's processing. This allows the model to make predictions using a
different value of subspace $\text{var}_i$ after a successful intervention.

DAS uses \emph{counterfactual behavior} to create intervention data to
train and evaluate $Q$.
Counterfactual behavior is defined as the behavior that \emph{would} have
occurred had the value of a variable in a causal abstraction been different and everything
else remained the same \citep{pearl2010causalmediation,geiger2021}.
We can create intervention data by freezing the state of an environment
at a particular point in time, changing one or more values in the causal
abstraction's variables, and then unfreezing the environment and using
the causal abstraction to generate the counterfactual behavior.
These intervention samples can then be used as training labels to
train $Q$ while keeping the NN parameters frozen. Once $Q$'s
training has converged, the robustness of the alignment can be evaluated
using the Interchange Intervention Accuracy (IIA)---defined as the
NN's prediction accuracy on the counterfactual behavior for a held out
set of interventions. IIA is then used to make claims about the NN's
internal mechanisms.

\section{Methods}
All of the analyses in this work are performed on autoregressive, sequence based ANNs
trained to predict sequences of tokens. Depending on the experiment, we consider
GRUs \citep{Cho2014GRU}, LSTMs \citep{hochreiter1997lstm}, shallow transformers \citep{vaswani2017,su2023roformer}, and DeepSeek-R1-Distill-Qwen-1.5B \citep{guo2025deepseek}.
The majority of our analyses are confined to models trained to perform variations
of numeric tasks. These tasks serve as a simplified setting
to demonstrate how MAS works and how it compares to other methods.
We additionally include a more general language modeling experiment where
we explore the similarity of toxic vs nontoxic models as a case study on
how to use MAS in more practical settings (Appendix~\ref{sup:toxicity}).

\subsection{Numeric Tasks}
Each task consists of a sequence of tokens that start with a
beginning of sequence token, B, and end with an end of sequence
token, E. Some of the tokens in each task are produced by the task
environment and define the specific goals within the task.
Other tokens are determined by these task provided tokens.
Each trial is considered correct when all deterministic tokens
are correctly predicted. During the model training, we include all
token types in a Next-Token Prediction (NTP) cross entropy loss, even though many tokens are
unpredictable. See Figure~\ref{fig:clmastaskcomp}(c) for a visual depiction of each of the tasks described in this section.

\textbf{\multiobject\space Task:} The environment presents some number
of demonstration (demo) tokens that are each sampled with replacement
from the set \{$\text{D}_a$, $\text{D}_b$, $\text{D}_c$\}. The task
is to produce the same number of response (R) tokens as D tokens and
end with an E token. The environment signals the end of the D
tokens by producing a trigger (T) token. The number of D tokens at
this point is referred to as the \emph{object quantity} for the trial,
which is uniformly sampled from 1 to \maxcount\space at the beginning.
The set of possible tokens includes \{B, $\text{D}_a$, $\text{D}_b$,
$\text{D}_c$, T, R, E\}. An example sequence with an object quantity of
2 is: "B $\text{D}_c$ $\text{D}_a$ T R R E"

To prevent solutions that use positional readout \citep{grant2024das},
we modify the task for transformer trainings: each token in the
demo phase has a \varyprob\space probability of being a
"void" token type, V, that has no impact on the object quantity. An example
sequence with an object count of 2 could be:
"B V $\text{D}_b$ V V $\text{D}_c$ T R R E". All evaluations and analyses
use the original \multiobject\space Task.

\textbf{\sameobject\space Task:} same structure as the \multiobject\space
task except that all D and R tokens are replaced by a single token type,
C. The set of possible tokens includes \{B, C, T, E\}. An example sequence
with an object quantity of 2 would be: "B C C T C C E".

\textbf{\modulo:}
This task is similar to the \multiobject\space task except
the number of R tokens is equal to the object
quantity mod \modulus. An example trial could be,
"B $\text{D}_b$ $\text{D}_c$ $\text{D}_a$ $\text{D}_c$ $\text{D}_b$ T R E".

\textbf{\rounding:}
Similar to the \modulo\space task except
the number of R tokens is equal to the object
quantity rounded to the nearest multiple of \roundmultiple. An example trial
could be, "B $\text{D}_b$ $\text{D}_c$ $\text{D}_a$ $\text{D}_c$ T R R R E".

\subsection{Model Architectures}
In our numeric task experiments, each model, $\psi$, is autoregressively trained to perform
only one of the tasks through next-token prediction (NTP). We train
\nseeds\space model seeds for each task variant up to $>99.99\%$ accuracy
on both training and validation data and freeze the weights before analysis
and interpretation. We consider Gated Recurrent Units (GRUs)
\citep{Cho2014GRU}, Long-Short Term Memory recurrent networks (LSTMs)
\citep{hochreiter1997lstm}, and two layer Transformers based on the
Roformer architecture \citep{vaswani2017,touvron2023llama,su2023roformer}.
The custom GRUs and Transformers use a dimensionality of \dmodel, whereas the
LSTM uses \dmodelhalf\space dimensions for each the h and c vectors. We leave
details of GRU and LSTM cells to the referenced papers beyond noting 
that the GRU and LSTM based models have the structure:
\begin{align}
h_{t+1} &= g(h_t, x_t)\\
\hat{x}_{t+1} &= f(h_{t+1})
\end{align}
Where $h_t$ is the hidden state vector at step $t$, $x_t$ is the input token
at step $t$, $g$ is the recurrent function (either a GRU or LSTM cell), and
$f$ is a two layer (two matrix) feed-forward network (FFN) used to make a
prediction, $\hat{x}_{t+1}$, of the token at step $t+1$ from the updated
hidden state $h_{t+1}$.

The transformer architecture uses Rotary Positional Encodings (RoPE)
\citep{su2023roformer} and GELU nonlinearities \citep{hendrycks2023gelus}.
Transformers use a history of input tokens, $X_t = [x_1, x_2, ..., x_t]$,
at each time step, $t$, to make a prediction: $\hat{h}_{t} = g(X_t)$, and
$\hat{x}_{t+1} = f(h_t)$ where $g$ and $f$ are now sets of transformer
layers. We show results from 2 layer, single attention head transformers.
We refer readers to Figure~\ref{fig:tformerarch} and Appendix~\ref{sup:models} for more details.
For our experiments on DeepSeek-R1-Distill-Qwen-1.5B, we
refer readers to \cite{guo2025deepseek,yang2025qwen3} for architectural
details and Appendix~\ref{sup:toxicity} for
finetuning details.

\subsection{MAS Formulation}\label{sec:MASformulation}
At a high level, MAS can be thought of as a combination
of model stitching and DAS. MAS causally measures the degree
to which multiple models' behavioral subspaces align with each
other.

Using our notation from Sections~\ref{sec:stitchingformulation}
and ~\ref{sec:DASformulation}, MAS learns an invertible alignment function,
$\mathcal{A}_{\psi_i}: \mathbb{R}^{d_{\psi_i}} \rightarrow \mathbb{R}^{d_{\psi_i}}$
for each $\psi_i$ of $\mathbb{N}$ models in the analysis,
such that $\mathcal{A}_{\psi_i}(h_{\psi_i}) = z_{\psi_i}$, and within each
$z_{\psi_i}$, MAS performs interchange interventions to examine the degree
to which the subspaces are causally interchangeable for all included models. It does this
using both between and within model interchange interventions.
Effectively, MAS asks, does $\Vec{z}_{\psi_i, \text{var}_k} =
\Vec{z}_{\psi_j, \text{var}_k} \forall \, (i \in \mathbb{N},j\in \mathbb{N})$? We
limit our empirical analyses to $\mathbb{N}=2$ and we only
examine cases
where $\mathcal{A}_{\psi_i} = Q_{\psi_i} = a_{\psi_i}U_{\psi_i}$
where $a_{\psi_i}$ is a scalar and $U_{\psi_i} \in R^{d_{\psi_i}\times d_{\psi_i}}$
is an orthogonal matrix.

With these observations, we can generalize Equation~\ref{eq:interchange} to
the multi-model case for an intervention on the subspace for $\text{var}_k$:
\begin{equation}\label{eq:masinterchange}
    h_{\psi_i}^{(v)} = Q_{\psi_i}^{-1}((\mathcal{I}-D_{\psi_i,\text{var}_k})Q_{\psi_i}h^{(trg)}_{\psi_i} +
    D_{\psi_j,\text{var}_k}Q_{\psi_j}h^{(src)}_{\psi_j})
\end{equation}
Where $i$ and $j$ can be any model index in the set of all
models considered,
$Q_{\psi_i}$ is a scaled orthogonal rotation matrix for $\psi_i$,
$D_{\psi_i,\text{var}_k}\in R^{d_{\psi_i}\times d_{\psi_i}}$ is a diagonal, binary matrix, specific to each model,
with $d_{\text{var}_k}$ non-zero elements used to isolate the dimensions
corresponding to $\Vec{z}_{\text{var}_k}$, $\mathcal{I}\in R^{d_{\psi_i}\times d_{\psi_i}}$ is the identity matrix,
$h^{(src)}_{\psi_j}$ is the \emph{source vector} from which the subspace
is harvested, $h^{(trg)}_{\psi_i}$ is the \emph{target vector} into which
activity is substituted, and $h^{(v)}_{\psi_i}$ is the resulting intervened
vector that replaces $h^{(trg)}_{\psi_i}$ in $\psi^{(trg)}_i$'s
processing, allowing the model to make causally intervened predictions.
In cases where $d_{\psi_i} > d_{\psi_j}$, $D_{\psi_j}$ can be expanded
to shape $\mathbb{R}^{d_{\psi_i}\times d_{\psi_j}}$ using zero padding
or reduced by removing rows from the bottom when
$d_{\psi_i} < d_{\psi_j}$.
See Figure~\ref{fig:masintervention}(a) for a visualization.

For many of our analyses, we will perform MAS without segregating the
behaviorally relevant subspace. Concretely, this can be formalized as a
causal abstraction that uses a single \varfull\space variable in which all behaviorally
relevant information is encoded, $\Vec{z}_{full}$, and all extraneous
information is encoded in
$\Vec{z}_{extra}$, the behavioral null-space and dormant subspaces. For completion:
$z_{\psi_i} = \begin{bmatrix} \Vec{z}_{\psi_i,full} \\ \Vec{z}_{\psi_i,extra} \end{bmatrix}$,
and we will freely isolate and manipulate $\Vec{z}_{\psi_i,full}$ using
Equation~\ref{eq:interchange} or ~\ref{eq:masinterchange}.

\subsection{MAS Training}\label{sec:MAStraining}
The MAS training procedure is similar to DAS in that the training loss
is created using the output distribution from $\psi_i^{(trg)}$ conditioned on $h^{(v)}_{\psi_i}$ as predictions and
the counterfactual behavior of the causal abstraction as labels. In the case of
$\Vec{z}_{full}$, the counterfactual behavior is the same as
that of model stitching from Equation~\ref{eq:mstitchloss}, where the
training labels are the behavior from $\psi_{src}$ following the
intervention. Using $y$ to denote the counterfactual behavior
labels in the loss for a single intervention, the loss for a single intervention direction is as follows:
\begin{equation} \label{eq:masloss}
\mathcal{L}(Q^{(trg)}_{\psi_i}, Q^{(src)}_{\psi_j}) \;=\;
\mathbb{E}_{x,y \sim \mathcal{D}} \big[ \ell\big( f_{\psi_i}(h^{(v)}_{\psi_i}) \,\|\, y \big) \big]
\end{equation}
In Sections~\ref{sec:unimas} and ~\ref{res:xmas} we will address cases in which we only include a subset
of the intervention direction permutations in the training loss.
However, unless otherwise stated,
we default to averaging over all permutations. The total loss for a single
matrix $Q_{\psi_i}$ is as follows:
\begin{equation}\label{eq:mastotloss}
\mathcal{L}_{total}(Q_{\psi_i}) \;=\;
\frac{1}{2\mathbb{N}-1}\sum_{j=1}^{\mathbb{N}} \big( \mathcal{L}(Q_{\psi_i}, Q_{\psi_j}) +
 \mathbf{1}_{\{i\neq j\}}\mathcal{L}(Q_{\psi_j}, Q_{\psi_i})  \big)
\end{equation}
When using auto-differentiation frameworks \citep{pytorch2019}, we simply use the gradient from the mean of the losses in all directions:
\begin{equation}\label{eq:differentiationloss}
\mathcal{L}_{total} \;=\;
\frac{1}{\mathbb{N}^2} \sum_{i=1}^{\mathbb{N}} \sum_{j=1}^{\mathbb{N}} \mathcal{L}(Q_{\psi_i}, Q_{\psi_j})
\end{equation}
It is crucial to include interventions where $i = j$ as this adds a
soft constraint to the alignment that encourages the separation of
$\Vec{z}_{full}$ from $\Vec{z}_{extra}$. This applies gradient pressure
for the interventions to only use causally relevant information.

\textbf{MAS Evaluation:}
We can evaluate the quality of each $Q_{\psi_i}$ using the accuracy of
the models' predictions on held out counterfactual data following the
interventions. We always report the IIA of the worst
performing causal direction in the analysis. In the
numeric tasks experiments, a trial is considered correct
when all deterministic tokens are predicted correctly using the argmax over
logits. For the LLM toxicity experiments, we report token
prediction accuracy rather than trial accuracy. We calculate error bars as
Standard Error over unique model seed pairings. We include two
additional seeds for within training type comparisons in the numeric task
experiments, and three seeds for each of the DeepSeek finetunings.

\subsection{Mas Variants}\label{sec:unimas}
We explore a number of variations on the MAS procedure as a means
of demonstrating how to use MAS in different situations for different
purposes. We enumerate these variants in this section.

\textbf{Stepwise MAS}:
Up to this point, we have described a form of MAS that performs interventions
on latent vectors at individual time points in the sequence. We include an
exploration of a variant in which Equation~\ref{eq:masinterchange} is
applied at each time point from the beginning of the sequence up to some
time point $t$. For all embedding layer analyses, we use this variant. See Figure~\ref{fig:masintervention}(b) for a visual depiction.

\textbf{\fullxmas\space (\xmas)}:
An interesting use case for neural similarity techniques is the comparison
of ANNs with BNNs. An issue with causal comparisons of this nature is that we often have no causal access to the BNN in the analysis. We do, however, usually have neural recordings from the BNN.
Using a tilde to denote causally inaccessible models,
$\tilde{\psi}_{i}$, we can simulate such ANN-BNN comparisons by omitting causal interventions using $\tilde{\psi}_{i}$ as the target model from the training loss.
We will assume, however, that we can still read from the latent
vectors produced by $\tilde{\psi}_i$---analogous to a neural
recording. We adapt Equation~\ref{eq:mastotloss} for $\psi_k$, where $k\neq i$, to ignore the behavior of
$\tilde{\psi}_i$ as follows:
\begin{eqnarray}\label{eq:UniMAStotloss}
\mathcal{L}_{accessible}(Q_{\psi_k}) = \frac{1}{2\mathbb{N}-1}
\big(
    \mathcal{L}( Q^{(trg)}_{\psi_k}, Q^{(src)}_{\tilde{\psi}_i}) +
    \sum_{j=1}^\mathbb{N} \mathbf{1}_{ \{j \neq i \}}
    \big(
        \mathcal{L}(Q^{(trg)}_{\psi_k},Q^{(src)}_{\psi_j}) +
        \mathbf{1}_{\{j\neq k\}}\mathcal{L}(Q^{(trg)}_{\psi_j},Q^{(src)}_{\psi_k})
    \big)
\big)
\end{eqnarray}
And the loss for $\tilde{\psi}_i$ is
$
    \mathcal{L}_{inaccessible}(Q_{\tilde{\psi}_i}) =
    \frac{1}{2\mathbb{N}-1}
    \sum_{j=1}^\mathbb{N} \mathbf{1}_{ \{j \neq i\}}  \mathcal{L}( Q^{(trg)}_{\psi_j}, Q^{(src)}_{\tilde{\psi}_i})
$.

We will show in Section~\ref{res:xmas} that if we attempt to train both
$Q_{\tilde{\psi}_i}$ and $Q_{\psi_j} \forall \, j \in \mathbb{N}_{ j \neq i}$
using only the behavioral gradient from minimizing Equation~\ref{eq:UniMAStotloss},
the resulting $Q$'s do not achieve high IIA when evaluating causal
interventions that use $\tilde{\psi}_i$ as the target model. To address
this shortcoming, we introduce an auxiliary loss function to encourage
causal relevance for $\tilde{\psi}_i^{(trg)}$.
This auxiliary objective relies on what we will refer to as
\emph{Counterfactual Latent (CL) vectors}, which we define as latent
vectors that encode the values of the causal variables that we would
expect to exist in the intervened vector---$h^{(v)}_{\tilde{\psi}_i}$
from Equation~\ref{eq:interchange}---after an interchange intervention.
If we focus on the \varfull\space variable case of $\Vec{z}_{full}$, the CL vectors are
prerecorded representations that produce to the same behavior as the
counterfactual behavior training labels. We can obtain CL vectors by
searching through a dataset of prerecorded $h_{\tilde{\psi}_i}$ vectors
for cases that either have the correct variable values produced by the
causal abstraction or those that lead to the same behavior as the counterfactual
behavior.

As an example, if we have a causal abstraction with
variables $\text{v}ar_1$, $\text{v}ar_2$, and $\text{v}ar_{extra}$, and,
we expect $h^{(v)}_{\tilde{\psi}_i}$ to
have a value of $y$ for variable $\text{v}ar_{1}$ and $w$ for variable
$\text{v}ar_2$ after applying Equation~\ref{eq:masinterchange},
then a valid CL vector, $h^{(CL)}_{\tilde{\psi}_i}$,
would be a
pre-recorded representation in which the causal abstraction has labeled
$h^{(CL)}_{\tilde{\psi}_i}$ to have
the variable values: $\text{v}ar_1 = y$ and $\text{v}ar_2 = w$.

The
auxiliary loss $\mathcal{X}$ for a single intervention sample is composed
of an L2 loss and a cosine loss using CL vectors as the ground truth:
\begin{eqnarray}\label{eq:auxobjective}
    \mathcal{X}_{L2} &=&  \frac{1}{2}||h_{\tilde{\psi}_i}^{v} - h^{(CL)}_{\tilde{\psi}_i}||^2_2 \\
    \mathcal{X}_{cos} &=&  -
    \frac{1}{2} \frac{h^{v}_{\tilde{\psi}_i} \cdot h^{(CL)}_{\tilde{\psi}_i}}{||h_{\tilde{\psi}_i}||_2\,||h^{(CL)}_{\tilde{\psi}_i}||_2}
\end{eqnarray}
where $h^{(v)}_{\tilde{\psi}_i}$ is the intervened target vector for the
causally inaccessible model. The total \xmas\space training loss is a weighted
sum of the loss from Equation~\ref{eq:UniMAStotloss} and the auxiliary loss
where $\epsilon$ is a hyperparameter:
$\mathcal{L}_{CL} = \epsilon(\mathcal{X}_{L2} + \mathcal{X}_{cos}) + (1-\epsilon)\mathcal{L}_{accessible}$. We show results from the
best performing training of $\epsilon$ values 0.5, 0.89, 0.94, and the best $d_{full}$ out of 32, 64, and 128 dimensions.


\textbf{Transferring Specific Variables}\label{sec:variables}
Through our choice of counterfactual training behavior, we can explore
alignment of specific subspaces between models that do not possess the
same domain or co-domain. We focus on representations of number as a case
study for their precise but general nature. To do this, we narrow the MAS interchange interventions to
$\Vec{z}_{\psi_1,numeric}$ and $\Vec{z}_{\psi_2,numeric}$ by constructing
counterfactual sequences for each $\psi$'s co-domain.
We assume that the models encode a single
numeric variable in the \multiobject, \sameobject, \modulo, and
\rounding\space tasks \cite{grant2024das}. In the arithmetic task, we
assume there is a numeric representation
for the number of remaining operations, \textbf{\varremops}, and another
representation for the cumulative value, \textbf{\varcumuval}. See
expanded details in Appendix~\ref{sup:variables}.

\textbf{Model Stitching:} As baselines, we include causal intervention methods
that we refer to as \textbf{Latent Stitch} and \textbf{Stitch}.
Latent Stitch consists of a single
orthogonal matrix $Q$ for two models, $\psi_1$ and $\psi_2$, that learns
to map $h^{(src)}_{\psi_1}$ to $h^{(CL)}_{\psi_2}$ by minimizing the CL
auxiliary loss from Equation~\ref{eq:auxobjective} without including the behavioral objective
Equation~\ref{eq:UniMAStotloss}. Stitch consists of a single, possibly
low rank, orthogonal matrix $Q$ trained in a single behavioral direction
from $\tilde{\psi}_1$ to $\psi_2$ without using the CL auxiliary loss.
Unless otherwise specified, assume $Q$ is full rank and the IIA is reported
from validation intervention data in the $\tilde{\psi}_1$ to $\psi_2$ direction only.

\newcommand{\GRUGRUEmbeddingsInputValCKAMean}{0.98}
\newcommand{\GRUGRUEmbeddingsInputValCKAError}{0.0}
\newcommand{\GRUGRUEmbeddingsInputValDASMean}{0.885}
\newcommand{\GRUGRUEmbeddingsInputValDASError}{0.04}
\newcommand{\GRUGRUEmbeddingsInputValMASMean}{0.854}
\newcommand{\GRUGRUEmbeddingsInputValMASError}{0.01}
\newcommand{\GRUGRUEmbeddingsInputValRSAMean}{0.7}
\newcommand{\GRUGRUEmbeddingsInputValRSAError}{0.0}
\newcommand{\GRUGRUHidStateCountDASMean}{0.9835}
\newcommand{\GRUGRUHidStateCountDASError}{0.008}
\newcommand{\GRUGRUHidStateCountMASMean}{0.98}
\newcommand{\GRUGRUHidStateCountMASError}{0.0}
\newcommand{\GRUGRUHidStateFullCKAMean}{0.99}
\newcommand{\GRUGRUHidStateFullCKAError}{0.0}
\newcommand{\GRUGRUHidStateFullMASMean}{0.986}
\newcommand{\GRUGRUHidStateFullMASError}{0.002}
\newcommand{\GRUGRUHidStateFullRSAMean}{0.98}
\newcommand{\GRUGRUHidStateFullRSAError}{0.0}
\newcommand{\GRUGRUSameObjEmbeddingsInputValCKAMean}{0.0921}
\newcommand{\GRUGRUSameObjEmbeddingsInputValCKAError}{0.005}
\newcommand{\GRUGRUSameObjEmbeddingsInputValDASMean}{0.885}
\newcommand{\GRUGRUSameObjEmbeddingsInputValDASError}{0.02}
\newcommand{\GRUGRUSameObjEmbeddingsInputValMASMean}{0.1995}
\newcommand{\GRUGRUSameObjEmbeddingsInputValMASError}{0.007}
\newcommand{\GRUGRUSameObjEmbeddingsInputValRSAMean}{0.054}
\newcommand{\GRUGRUSameObjEmbeddingsInputValRSAError}{0.02}
\newcommand{\GRUGRUSameObjHidStateCountDASMean}{0.9835}
\newcommand{\GRUGRUSameObjHidStateCountDASError}{0.004}
\newcommand{\GRUGRUSameObjHidStateCountMASMean}{0.4935}
\newcommand{\GRUGRUSameObjHidStateCountMASError}{0.009}
\newcommand{\GRUGRUSameObjHidStateFullCKAMean}{0.782}
\newcommand{\GRUGRUSameObjHidStateFullCKAError}{0.01}
\newcommand{\GRUGRUSameObjHidStateFullMASMean}{0.948}
\newcommand{\GRUGRUSameObjHidStateFullMASError}{0.01}
\newcommand{\GRUGRUSameObjHidStateFullRSAMean}{0.7504}
\newcommand{\GRUGRUSameObjHidStateFullRSAError}{0.008}
\newcommand{\GRULSTMEmbeddingsInputValCKAMean}{0.7445}
\newcommand{\GRULSTMEmbeddingsInputValCKAError}{0.006}
\newcommand{\GRULSTMEmbeddingsInputValDASMean}{0.885}
\newcommand{\GRULSTMEmbeddingsInputValDASError}{0.02}
\newcommand{\GRULSTMEmbeddingsInputValMASMean}{0.88}
\newcommand{\GRULSTMEmbeddingsInputValMASError}{0.01}
\newcommand{\GRULSTMEmbeddingsInputValRSAMean}{0.46}
\newcommand{\GRULSTMEmbeddingsInputValRSAError}{0.02}
\newcommand{\GRULSTMHidStateCountDASMean}{0.9835}
\newcommand{\GRULSTMHidStateCountDASError}{0.004}
\newcommand{\GRULSTMHidStateCountMASMean}{0.754}
\newcommand{\GRULSTMHidStateCountMASError}{0.05}
\newcommand{\GRULSTMHidStateFullCKAMean}{0.9326}
\newcommand{\GRULSTMHidStateFullCKAError}{0.003}
\newcommand{\GRULSTMHidStateFullMASMean}{0.797}
\newcommand{\GRULSTMHidStateFullMASError}{0.03}
\newcommand{\GRULSTMHidStateFullRSAMean}{0.9052}
\newcommand{\GRULSTMHidStateFullRSAError}{0.002}
\newcommand{\GRULSTMSameObjEmbeddingsInputValCKAMean}{0.0863}
\newcommand{\GRULSTMSameObjEmbeddingsInputValCKAError}{0.005}
\newcommand{\GRULSTMSameObjEmbeddingsInputValMASMean}{0.432}
\newcommand{\GRULSTMSameObjEmbeddingsInputValMASError}{0.01}
\newcommand{\GRULSTMSameObjEmbeddingsInputValRSAMean}{0.049}
\newcommand{\GRULSTMSameObjEmbeddingsInputValRSAError}{0.02}
\newcommand{\GRULSTMSameObjHidStateCountMASMean}{0.644}
\newcommand{\GRULSTMSameObjHidStateCountMASError}{0.07}
\newcommand{\GRULSTMSameObjHidStateFullCKAMean}{0.67}
\newcommand{\GRULSTMSameObjHidStateFullCKAError}{0.1}
\newcommand{\GRULSTMSameObjHidStateFullMASMean}{0.832}
\newcommand{\GRULSTMSameObjHidStateFullMASError}{0.05}
\newcommand{\GRULSTMSameObjHidStateFullRSAMean}{0.665}
\newcommand{\GRULSTMSameObjHidStateFullRSAError}{0.09}
\newcommand{\GRUTransformerEmbeddingsInputValCKAMean}{0.7808}
\newcommand{\GRUTransformerEmbeddingsInputValCKAError}{0.007}
\newcommand{\GRUTransformerEmbeddingsInputValDASMean}{0.885}
\newcommand{\GRUTransformerEmbeddingsInputValDASError}{0.02}
\newcommand{\GRUTransformerEmbeddingsInputValMASMean}{0.903}
\newcommand{\GRUTransformerEmbeddingsInputValMASError}{0.02}
\newcommand{\GRUTransformerEmbeddingsInputValRSAMean}{0.547}
\newcommand{\GRUTransformerEmbeddingsInputValRSAError}{0.04}
\newcommand{\GRUTransformerHidStateCountDASMean}{0.9835}
\newcommand{\GRUTransformerHidStateCountDASError}{0.004}
\newcommand{\GRUTransformerHidStateCountMASMean}{0.1108}
\newcommand{\GRUTransformerHidStateCountMASError}{0.008}
\newcommand{\GRUTransformerHidStateFullCKAMean}{0.8471}
\newcommand{\GRUTransformerHidStateFullCKAError}{0.006}
\newcommand{\GRUTransformerHidStateFullMASMean}{0.171}
\newcommand{\GRUTransformerHidStateFullMASError}{0.01}
\newcommand{\GRUTransformerHidStateFullRSAMean}{0.755}
\newcommand{\GRUTransformerHidStateFullRSAError}{0.03}
\newcommand{\GRUSameObjGRUEmbeddingsInputValCKAMean}{0.0921}
\newcommand{\GRUSameObjGRUEmbeddingsInputValCKAError}{0.005}
\newcommand{\GRUSameObjGRUEmbeddingsInputValMASMean}{0.1995}
\newcommand{\GRUSameObjGRUEmbeddingsInputValMASError}{0.007}
\newcommand{\GRUSameObjGRUEmbeddingsInputValRSAMean}{0.054}
\newcommand{\GRUSameObjGRUEmbeddingsInputValRSAError}{0.02}
\newcommand{\GRUSameObjGRUHidStateCountMASMean}{0.4935}
\newcommand{\GRUSameObjGRUHidStateCountMASError}{0.009}
\newcommand{\GRUSameObjGRUHidStateFullCKAMean}{0.782}
\newcommand{\GRUSameObjGRUHidStateFullCKAError}{0.01}
\newcommand{\GRUSameObjGRUHidStateFullMASMean}{0.948}
\newcommand{\GRUSameObjGRUHidStateFullMASError}{0.01}
\newcommand{\GRUSameObjGRUHidStateFullRSAMean}{0.7504}
\newcommand{\GRUSameObjGRUHidStateFullRSAError}{0.008}
\newcommand{\GRUSameObjGRUSameObjEmbeddingsInputValCKAMean}{0.9}
\newcommand{\GRUSameObjGRUSameObjEmbeddingsInputValCKAError}{0.0}
\newcommand{\GRUSameObjGRUSameObjEmbeddingsInputValDASMean}{0.4545}
\newcommand{\GRUSameObjGRUSameObjEmbeddingsInputValDASError}{0.004}
\newcommand{\GRUSameObjGRUSameObjEmbeddingsInputValMASMean}{0.1595}
\newcommand{\GRUSameObjGRUSameObjEmbeddingsInputValMASError}{0.009}
\newcommand{\GRUSameObjGRUSameObjEmbeddingsInputValRSAMean}{0.9}
\newcommand{\GRUSameObjGRUSameObjEmbeddingsInputValRSAError}{0.0}
\newcommand{\GRUSameObjGRUSameObjHidStateCountDASMean}{0.4165}
\newcommand{\GRUSameObjGRUSameObjHidStateCountDASError}{0.002}
\newcommand{\GRUSameObjGRUSameObjHidStateCountMASMean}{0.4665}
\newcommand{\GRUSameObjGRUSameObjHidStateCountMASError}{0.0004}
\newcommand{\GRUSameObjGRUSameObjHidStateFullCKAMean}{0.96}
\newcommand{\GRUSameObjGRUSameObjHidStateFullCKAError}{0.0}
\newcommand{\GRUSameObjGRUSameObjHidStateFullMASMean}{0.981}
\newcommand{\GRUSameObjGRUSameObjHidStateFullMASError}{0.001}
\newcommand{\GRUSameObjGRUSameObjHidStateFullRSAMean}{0.98}
\newcommand{\GRUSameObjGRUSameObjHidStateFullRSAError}{0.0}
\newcommand{\GRUSameObjLSTMEmbeddingsInputValCKAMean}{0.0488}
\newcommand{\GRUSameObjLSTMEmbeddingsInputValCKAError}{0.002}
\newcommand{\GRUSameObjLSTMEmbeddingsInputValMASMean}{0.397}
\newcommand{\GRUSameObjLSTMEmbeddingsInputValMASError}{0.03}
\newcommand{\GRUSameObjLSTMEmbeddingsInputValRSAMean}{0.078}
\newcommand{\GRUSameObjLSTMEmbeddingsInputValRSAError}{0.03}
\newcommand{\GRUSameObjLSTMHidStateCountMASMean}{0.489}
\newcommand{\GRUSameObjLSTMHidStateCountMASError}{0.01}
\newcommand{\GRUSameObjLSTMHidStateFullCKAMean}{0.75}
\newcommand{\GRUSameObjLSTMHidStateFullCKAError}{0.01}
\newcommand{\GRUSameObjLSTMHidStateFullMASMean}{0.85}
\newcommand{\GRUSameObjLSTMHidStateFullMASError}{0.05}
\newcommand{\GRUSameObjLSTMHidStateFullRSAMean}{0.677}
\newcommand{\GRUSameObjLSTMHidStateFullRSAError}{0.003}
\newcommand{\GRUSameObjLSTMSameObjEmbeddingsInputValCKAMean}{0.472}
\newcommand{\GRUSameObjLSTMSameObjEmbeddingsInputValCKAError}{0.04}
\newcommand{\GRUSameObjLSTMSameObjEmbeddingsInputValDASMean}{0.4545}
\newcommand{\GRUSameObjLSTMSameObjEmbeddingsInputValDASError}{0.002}
\newcommand{\GRUSameObjLSTMSameObjEmbeddingsInputValMASMean}{0.251}
\newcommand{\GRUSameObjLSTMSameObjEmbeddingsInputValMASError}{0.01}
\newcommand{\GRUSameObjLSTMSameObjEmbeddingsInputValRSAMean}{0.635}
\newcommand{\GRUSameObjLSTMSameObjEmbeddingsInputValRSAError}{0.03}
\newcommand{\GRUSameObjLSTMSameObjHidStateCountDASMean}{0.4165}
\newcommand{\GRUSameObjLSTMSameObjHidStateCountDASError}{0.001}
\newcommand{\GRUSameObjLSTMSameObjHidStateCountMASMean}{0.4995}
\newcommand{\GRUSameObjLSTMSameObjHidStateCountMASError}{0.004}
\newcommand{\GRUSameObjLSTMSameObjHidStateFullCKAMean}{0.669}
\newcommand{\GRUSameObjLSTMSameObjHidStateFullCKAError}{0.04}
\newcommand{\GRUSameObjLSTMSameObjHidStateFullMASMean}{0.79}
\newcommand{\GRUSameObjLSTMSameObjHidStateFullMASError}{0.1}
\newcommand{\GRUSameObjLSTMSameObjHidStateFullRSAMean}{0.685}
\newcommand{\GRUSameObjLSTMSameObjHidStateFullRSAError}{0.07}
\newcommand{\GRUSameObjTransformerEmbeddingsInputValCKAMean}{0.0525}
\newcommand{\GRUSameObjTransformerEmbeddingsInputValCKAError}{0.002}
\newcommand{\GRUSameObjTransformerEmbeddingsInputValMASMean}{0.2028}
\newcommand{\GRUSameObjTransformerEmbeddingsInputValMASError}{0.002}
\newcommand{\GRUSameObjTransformerEmbeddingsInputValRSAMean}{0.176}
\newcommand{\GRUSameObjTransformerEmbeddingsInputValRSAError}{0.01}
\newcommand{\GRUSameObjTransformerHidStateCountMASMean}{0.1013}
\newcommand{\GRUSameObjTransformerHidStateCountMASError}{0.007}
\newcommand{\GRUSameObjTransformerHidStateFullCKAMean}{0.63}
\newcommand{\GRUSameObjTransformerHidStateFullCKAError}{0.01}
\newcommand{\GRUSameObjTransformerHidStateFullMASMean}{0.166}
\newcommand{\GRUSameObjTransformerHidStateFullMASError}{0.008}
\newcommand{\GRUSameObjTransformerHidStateFullRSAMean}{0.499}
\newcommand{\GRUSameObjTransformerHidStateFullRSAError}{0.02}
\newcommand{\LSTMGRUEmbeddingsInputValCKAMean}{0.7445}
\newcommand{\LSTMGRUEmbeddingsInputValCKAError}{0.006}
\newcommand{\LSTMGRUEmbeddingsInputValDASMean}{0.934}
\newcommand{\LSTMGRUEmbeddingsInputValDASError}{0.006}
\newcommand{\LSTMGRUEmbeddingsInputValMASMean}{0.88}
\newcommand{\LSTMGRUEmbeddingsInputValMASError}{0.01}
\newcommand{\LSTMGRUEmbeddingsInputValRSAMean}{0.46}
\newcommand{\LSTMGRUEmbeddingsInputValRSAError}{0.02}
\newcommand{\LSTMGRUHidStateCountDASMean}{0.9875}
\newcommand{\LSTMGRUHidStateCountDASError}{0.002}
\newcommand{\LSTMGRUHidStateCountMASMean}{0.754}
\newcommand{\LSTMGRUHidStateCountMASError}{0.05}
\newcommand{\LSTMGRUHidStateFullCKAMean}{0.9326}
\newcommand{\LSTMGRUHidStateFullCKAError}{0.003}
\newcommand{\LSTMGRUHidStateFullMASMean}{0.797}
\newcommand{\LSTMGRUHidStateFullMASError}{0.03}
\newcommand{\LSTMGRUHidStateFullRSAMean}{0.9052}
\newcommand{\LSTMGRUHidStateFullRSAError}{0.002}
\newcommand{\LSTMGRUSameObjEmbeddingsInputValCKAMean}{0.0488}
\newcommand{\LSTMGRUSameObjEmbeddingsInputValCKAError}{0.002}
\newcommand{\LSTMGRUSameObjEmbeddingsInputValDASMean}{0.934}
\newcommand{\LSTMGRUSameObjEmbeddingsInputValDASError}{0.006}
\newcommand{\LSTMGRUSameObjEmbeddingsInputValMASMean}{0.397}
\newcommand{\LSTMGRUSameObjEmbeddingsInputValMASError}{0.03}
\newcommand{\LSTMGRUSameObjEmbeddingsInputValRSAMean}{0.078}
\newcommand{\LSTMGRUSameObjEmbeddingsInputValRSAError}{0.03}
\newcommand{\LSTMGRUSameObjHidStateCountDASMean}{0.9875}
\newcommand{\LSTMGRUSameObjHidStateCountDASError}{0.002}
\newcommand{\LSTMGRUSameObjHidStateCountMASMean}{0.489}
\newcommand{\LSTMGRUSameObjHidStateCountMASError}{0.01}
\newcommand{\LSTMGRUSameObjHidStateFullCKAMean}{0.75}
\newcommand{\LSTMGRUSameObjHidStateFullCKAError}{0.01}
\newcommand{\LSTMGRUSameObjHidStateFullMASMean}{0.85}
\newcommand{\LSTMGRUSameObjHidStateFullMASError}{0.05}
\newcommand{\LSTMGRUSameObjHidStateFullRSAMean}{0.677}
\newcommand{\LSTMGRUSameObjHidStateFullRSAError}{0.003}
\newcommand{\LSTMLSTMEmbeddingsInputValCKAMean}{0.96}
\newcommand{\LSTMLSTMEmbeddingsInputValCKAError}{0.0}
\newcommand{\LSTMLSTMEmbeddingsInputValDASMean}{0.934}
\newcommand{\LSTMLSTMEmbeddingsInputValDASError}{0.01}
\newcommand{\LSTMLSTMEmbeddingsInputValMASMean}{0.915}
\newcommand{\LSTMLSTMEmbeddingsInputValMASError}{0.002}
\newcommand{\LSTMLSTMEmbeddingsInputValRSAMean}{0.55}
\newcommand{\LSTMLSTMEmbeddingsInputValRSAError}{0.0}
\newcommand{\LSTMLSTMHidStateCountDASMean}{0.9875}
\newcommand{\LSTMLSTMHidStateCountDASError}{0.003}
\newcommand{\LSTMLSTMHidStateCountMASMean}{0.986}
\newcommand{\LSTMLSTMHidStateCountMASError}{0.004}
\newcommand{\LSTMLSTMHidStateFullCKAMean}{0.99}
\newcommand{\LSTMLSTMHidStateFullCKAError}{0.0}
\newcommand{\LSTMLSTMHidStateFullMASMean}{0.986}
\newcommand{\LSTMLSTMHidStateFullMASError}{0.002}
\newcommand{\LSTMLSTMHidStateFullRSAMean}{0.99}
\newcommand{\LSTMLSTMHidStateFullRSAError}{0.0}
\newcommand{\LSTMLSTMSameObjEmbeddingsInputValCKAMean}{0.105}
\newcommand{\LSTMLSTMSameObjEmbeddingsInputValCKAError}{0.01}
\newcommand{\LSTMLSTMSameObjEmbeddingsInputValDASMean}{0.934}
\newcommand{\LSTMLSTMSameObjEmbeddingsInputValDASError}{0.006}
\newcommand{\LSTMLSTMSameObjEmbeddingsInputValMASMean}{0.307}
\newcommand{\LSTMLSTMSameObjEmbeddingsInputValMASError}{0.03}
\newcommand{\LSTMLSTMSameObjEmbeddingsInputValRSAMean}{0.087}
\newcommand{\LSTMLSTMSameObjEmbeddingsInputValRSAError}{0.06}
\newcommand{\LSTMLSTMSameObjHidStateCountMASMean}{0.73}
\newcommand{\LSTMLSTMSameObjHidStateCountMASError}{0.1}
\newcommand{\LSTMLSTMSameObjHidStateFullCKAMean}{0.72}
\newcommand{\LSTMLSTMSameObjHidStateFullCKAError}{0.1}
\newcommand{\LSTMLSTMSameObjHidStateFullMASMean}{0.847}
\newcommand{\LSTMLSTMSameObjHidStateFullMASError}{0.08}
\newcommand{\LSTMLSTMSameObjHidStateFullRSAMean}{0.71}
\newcommand{\LSTMLSTMSameObjHidStateFullRSAError}{0.1}
\newcommand{\LSTMTransformerEmbeddingsInputValCKAMean}{0.9663}
\newcommand{\LSTMTransformerEmbeddingsInputValCKAError}{0.007}
\newcommand{\LSTMTransformerEmbeddingsInputValDASMean}{0.934}
\newcommand{\LSTMTransformerEmbeddingsInputValDASError}{0.006}
\newcommand{\LSTMTransformerEmbeddingsInputValMASMean}{0.914}
\newcommand{\LSTMTransformerEmbeddingsInputValMASError}{0.006}
\newcommand{\LSTMTransformerEmbeddingsInputValRSAMean}{0.73}
\newcommand{\LSTMTransformerEmbeddingsInputValRSAError}{0.08}
\newcommand{\LSTMTransformerHidStateCountDASMean}{0.9875}
\newcommand{\LSTMTransformerHidStateCountDASError}{0.002}
\newcommand{\LSTMTransformerHidStateCountMASMean}{0.1065}
\newcommand{\LSTMTransformerHidStateCountMASError}{0.007}
\newcommand{\LSTMTransformerHidStateFullCKAMean}{0.804}
\newcommand{\LSTMTransformerHidStateFullCKAError}{0.008}
\newcommand{\LSTMTransformerHidStateFullMASMean}{0.159}
\newcommand{\LSTMTransformerHidStateFullMASError}{0.005}
\newcommand{\LSTMTransformerHidStateFullRSAMean}{0.723}
\newcommand{\LSTMTransformerHidStateFullRSAError}{0.04}
\newcommand{\LSTMSameObjGRUEmbeddingsInputValCKAMean}{0.0863}
\newcommand{\LSTMSameObjGRUEmbeddingsInputValCKAError}{0.005}
\newcommand{\LSTMSameObjGRUEmbeddingsInputValMASMean}{0.432}
\newcommand{\LSTMSameObjGRUEmbeddingsInputValMASError}{0.01}
\newcommand{\LSTMSameObjGRUEmbeddingsInputValRSAMean}{0.049}
\newcommand{\LSTMSameObjGRUEmbeddingsInputValRSAError}{0.02}
\newcommand{\LSTMSameObjGRUHidStateCountMASMean}{0.644}
\newcommand{\LSTMSameObjGRUHidStateCountMASError}{0.07}
\newcommand{\LSTMSameObjGRUHidStateFullCKAMean}{0.67}
\newcommand{\LSTMSameObjGRUHidStateFullCKAError}{0.1}
\newcommand{\LSTMSameObjGRUHidStateFullMASMean}{0.832}
\newcommand{\LSTMSameObjGRUHidStateFullMASError}{0.05}
\newcommand{\LSTMSameObjGRUHidStateFullRSAMean}{0.665}
\newcommand{\LSTMSameObjGRUHidStateFullRSAError}{0.09}
\newcommand{\LSTMSameObjGRUSameObjEmbeddingsInputValCKAMean}{0.472}
\newcommand{\LSTMSameObjGRUSameObjEmbeddingsInputValCKAError}{0.04}
\newcommand{\LSTMSameObjGRUSameObjEmbeddingsInputValMASMean}{0.251}
\newcommand{\LSTMSameObjGRUSameObjEmbeddingsInputValMASError}{0.01}
\newcommand{\LSTMSameObjGRUSameObjEmbeddingsInputValRSAMean}{0.635}
\newcommand{\LSTMSameObjGRUSameObjEmbeddingsInputValRSAError}{0.03}
\newcommand{\LSTMSameObjGRUSameObjHidStateCountMASMean}{0.4995}
\newcommand{\LSTMSameObjGRUSameObjHidStateCountMASError}{0.004}
\newcommand{\LSTMSameObjGRUSameObjHidStateFullCKAMean}{0.669}
\newcommand{\LSTMSameObjGRUSameObjHidStateFullCKAError}{0.04}
\newcommand{\LSTMSameObjGRUSameObjHidStateFullMASMean}{0.79}
\newcommand{\LSTMSameObjGRUSameObjHidStateFullMASError}{0.1}
\newcommand{\LSTMSameObjGRUSameObjHidStateFullRSAMean}{0.685}
\newcommand{\LSTMSameObjGRUSameObjHidStateFullRSAError}{0.07}
\newcommand{\LSTMSameObjLSTMEmbeddingsInputValCKAMean}{0.105}
\newcommand{\LSTMSameObjLSTMEmbeddingsInputValCKAError}{0.01}
\newcommand{\LSTMSameObjLSTMEmbeddingsInputValMASMean}{0.324}
\newcommand{\LSTMSameObjLSTMEmbeddingsInputValMASError}{0.05}
\newcommand{\LSTMSameObjLSTMEmbeddingsInputValRSAMean}{0.087}
\newcommand{\LSTMSameObjLSTMEmbeddingsInputValRSAError}{0.06}
\newcommand{\LSTMSameObjLSTMHidStateCountMASMean}{0.73}
\newcommand{\LSTMSameObjLSTMHidStateCountMASError}{0.1}
\newcommand{\LSTMSameObjLSTMHidStateFullCKAMean}{0.72}
\newcommand{\LSTMSameObjLSTMHidStateFullCKAError}{0.1}
\newcommand{\LSTMSameObjLSTMHidStateFullMASMean}{0.847}
\newcommand{\LSTMSameObjLSTMHidStateFullMASError}{0.08}
\newcommand{\LSTMSameObjLSTMHidStateFullRSAMean}{0.71}
\newcommand{\LSTMSameObjLSTMHidStateFullRSAError}{0.1}
\newcommand{\LSTMSameObjLSTMSameObjEmbeddingsInputValCKAMean}{0.22}
\newcommand{\LSTMSameObjLSTMSameObjEmbeddingsInputValCKAError}{0.0}
\newcommand{\LSTMSameObjLSTMSameObjEmbeddingsInputValMASMean}{0.266}
\newcommand{\LSTMSameObjLSTMSameObjEmbeddingsInputValMASError}{0.003}
\newcommand{\LSTMSameObjLSTMSameObjEmbeddingsInputValRSAMean}{0.43}
\newcommand{\LSTMSameObjLSTMSameObjEmbeddingsInputValRSAError}{0.0}
\newcommand{\LSTMSameObjLSTMSameObjHidStateCountMASMean}{0.518}
\newcommand{\LSTMSameObjLSTMSameObjHidStateCountMASError}{0.01}
\newcommand{\LSTMSameObjLSTMSameObjHidStateFullCKAMean}{0.5}
\newcommand{\LSTMSameObjLSTMSameObjHidStateFullCKAError}{0.0}
\newcommand{\LSTMSameObjLSTMSameObjHidStateFullMASMean}{0.86}
\newcommand{\LSTMSameObjLSTMSameObjHidStateFullMASError}{0.1}
\newcommand{\LSTMSameObjLSTMSameObjHidStateFullRSAMean}{0.54}
\newcommand{\LSTMSameObjLSTMSameObjHidStateFullRSAError}{0.0}
\newcommand{\LSTMSameObjTransformerEmbeddingsInputValCKAMean}{0.1004}
\newcommand{\LSTMSameObjTransformerEmbeddingsInputValCKAError}{0.007}
\newcommand{\LSTMSameObjTransformerEmbeddingsInputValMASMean}{0.427}
\newcommand{\LSTMSameObjTransformerEmbeddingsInputValMASError}{0.01}
\newcommand{\LSTMSameObjTransformerEmbeddingsInputValRSAMean}{0.158}
\newcommand{\LSTMSameObjTransformerEmbeddingsInputValRSAError}{0.06}
\newcommand{\LSTMSameObjTransformerHidStateCountMASMean}{0.1138}
\newcommand{\LSTMSameObjTransformerHidStateCountMASError}{0.007}
\newcommand{\LSTMSameObjTransformerHidStateFullCKAMean}{0.54}
\newcommand{\LSTMSameObjTransformerHidStateFullCKAError}{0.1}
\newcommand{\LSTMSameObjTransformerHidStateFullMASMean}{0.167}
\newcommand{\LSTMSameObjTransformerHidStateFullMASError}{0.006}
\newcommand{\LSTMSameObjTransformerHidStateFullRSAMean}{0.466}
\newcommand{\LSTMSameObjTransformerHidStateFullRSAError}{0.07}
\newcommand{\TransformerGRUEmbeddingsInputValCKAMean}{0.7808}
\newcommand{\TransformerGRUEmbeddingsInputValCKAError}{0.007}
\newcommand{\TransformerGRUEmbeddingsInputValDASMean}{0.943}
\newcommand{\TransformerGRUEmbeddingsInputValDASError}{0.008}
\newcommand{\TransformerGRUEmbeddingsInputValMASMean}{0.903}
\newcommand{\TransformerGRUEmbeddingsInputValMASError}{0.02}
\newcommand{\TransformerGRUEmbeddingsInputValRSAMean}{0.547}
\newcommand{\TransformerGRUEmbeddingsInputValRSAError}{0.04}
\newcommand{\TransformerGRUHidStateCountDASMean}{0.113}
\newcommand{\TransformerGRUHidStateCountDASError}{0.006}
\newcommand{\TransformerGRUHidStateCountMASMean}{0.1108}
\newcommand{\TransformerGRUHidStateCountMASError}{0.008}
\newcommand{\TransformerGRUHidStateFullCKAMean}{0.8471}
\newcommand{\TransformerGRUHidStateFullCKAError}{0.006}
\newcommand{\TransformerGRUHidStateFullMASMean}{0.171}
\newcommand{\TransformerGRUHidStateFullMASError}{0.01}
\newcommand{\TransformerGRUHidStateFullRSAMean}{0.755}
\newcommand{\TransformerGRUHidStateFullRSAError}{0.03}
\newcommand{\TransformerGRUSameObjEmbeddingsInputValCKAMean}{0.0525}
\newcommand{\TransformerGRUSameObjEmbeddingsInputValCKAError}{0.002}
\newcommand{\TransformerGRUSameObjEmbeddingsInputValDASMean}{0.943}
\newcommand{\TransformerGRUSameObjEmbeddingsInputValDASError}{0.008}
\newcommand{\TransformerGRUSameObjEmbeddingsInputValMASMean}{0.2028}
\newcommand{\TransformerGRUSameObjEmbeddingsInputValMASError}{0.002}
\newcommand{\TransformerGRUSameObjEmbeddingsInputValRSAMean}{0.176}
\newcommand{\TransformerGRUSameObjEmbeddingsInputValRSAError}{0.01}
\newcommand{\TransformerGRUSameObjHidStateCountDASMean}{0.113}
\newcommand{\TransformerGRUSameObjHidStateCountDASError}{0.006}
\newcommand{\TransformerGRUSameObjHidStateCountMASMean}{0.1013}
\newcommand{\TransformerGRUSameObjHidStateCountMASError}{0.007}
\newcommand{\TransformerGRUSameObjHidStateFullCKAMean}{0.63}
\newcommand{\TransformerGRUSameObjHidStateFullCKAError}{0.01}
\newcommand{\TransformerGRUSameObjHidStateFullMASMean}{0.166}
\newcommand{\TransformerGRUSameObjHidStateFullMASError}{0.008}
\newcommand{\TransformerGRUSameObjHidStateFullRSAMean}{0.499}
\newcommand{\TransformerGRUSameObjHidStateFullRSAError}{0.02}
\newcommand{\TransformerLSTMEmbeddingsInputValCKAMean}{0.9663}
\newcommand{\TransformerLSTMEmbeddingsInputValCKAError}{0.007}
\newcommand{\TransformerLSTMEmbeddingsInputValDASMean}{0.943}
\newcommand{\TransformerLSTMEmbeddingsInputValDASError}{0.008}
\newcommand{\TransformerLSTMEmbeddingsInputValMASMean}{0.914}
\newcommand{\TransformerLSTMEmbeddingsInputValMASError}{0.006}
\newcommand{\TransformerLSTMEmbeddingsInputValRSAMean}{0.73}
\newcommand{\TransformerLSTMEmbeddingsInputValRSAError}{0.08}
\newcommand{\TransformerLSTMHidStateCountDASMean}{0.113}
\newcommand{\TransformerLSTMHidStateCountDASError}{0.006}
\newcommand{\TransformerLSTMHidStateCountMASMean}{0.1065}
\newcommand{\TransformerLSTMHidStateCountMASError}{0.007}
\newcommand{\TransformerLSTMHidStateFullCKAMean}{0.804}
\newcommand{\TransformerLSTMHidStateFullCKAError}{0.008}
\newcommand{\TransformerLSTMHidStateFullMASMean}{0.159}
\newcommand{\TransformerLSTMHidStateFullMASError}{0.005}
\newcommand{\TransformerLSTMHidStateFullRSAMean}{0.723}
\newcommand{\TransformerLSTMHidStateFullRSAError}{0.04}
\newcommand{\TransformerLSTMSameObjEmbeddingsInputValCKAMean}{0.1004}
\newcommand{\TransformerLSTMSameObjEmbeddingsInputValCKAError}{0.007}
\newcommand{\TransformerLSTMSameObjEmbeddingsInputValMASMean}{0.427}
\newcommand{\TransformerLSTMSameObjEmbeddingsInputValMASError}{0.01}
\newcommand{\TransformerLSTMSameObjEmbeddingsInputValRSAMean}{0.158}
\newcommand{\TransformerLSTMSameObjEmbeddingsInputValRSAError}{0.06}
\newcommand{\TransformerLSTMSameObjHidStateCountMASMean}{0.1138}
\newcommand{\TransformerLSTMSameObjHidStateCountMASError}{0.007}
\newcommand{\TransformerLSTMSameObjHidStateFullCKAMean}{0.54}
\newcommand{\TransformerLSTMSameObjHidStateFullCKAError}{0.1}
\newcommand{\TransformerLSTMSameObjHidStateFullMASMean}{0.167}
\newcommand{\TransformerLSTMSameObjHidStateFullMASError}{0.006}
\newcommand{\TransformerLSTMSameObjHidStateFullRSAMean}{0.466}
\newcommand{\TransformerLSTMSameObjHidStateFullRSAError}{0.07}
\newcommand{\TransformerTransformerEmbeddingsInputValCKAMean}{0.98}
\newcommand{\TransformerTransformerEmbeddingsInputValCKAError}{0.0}
\newcommand{\TransformerTransformerEmbeddingsInputValDASMean}{0.943}
\newcommand{\TransformerTransformerEmbeddingsInputValDASError}{0.01}
\newcommand{\TransformerTransformerEmbeddingsInputValMASMean}{0.961}
\newcommand{\TransformerTransformerEmbeddingsInputValMASError}{0.004}
\newcommand{\TransformerTransformerEmbeddingsInputValRSAMean}{0.88}
\newcommand{\TransformerTransformerEmbeddingsInputValRSAError}{0.0}
\newcommand{\TransformerTransformerHidStateCountDASMean}{0.113}
\newcommand{\TransformerTransformerHidStateCountDASError}{0.01}
\newcommand{\TransformerTransformerHidStateCountMASMean}{0.087}
\newcommand{\TransformerTransformerHidStateCountMASError}{0.01}
\newcommand{\TransformerTransformerHidStateFullCKAMean}{0.96}
\newcommand{\TransformerTransformerHidStateFullCKAError}{0.0}
\newcommand{\TransformerTransformerHidStateFullMASMean}{0.148}
\newcommand{\TransformerTransformerHidStateFullMASError}{0.004}
\newcommand{\TransformerTransformerHidStateFullRSAMean}{0.78}
\newcommand{\TransformerTransformerHidStateFullRSAError}{0.0}
\begin{figure}[th!]
    \centering
    \includegraphics[width=\columnwidth]{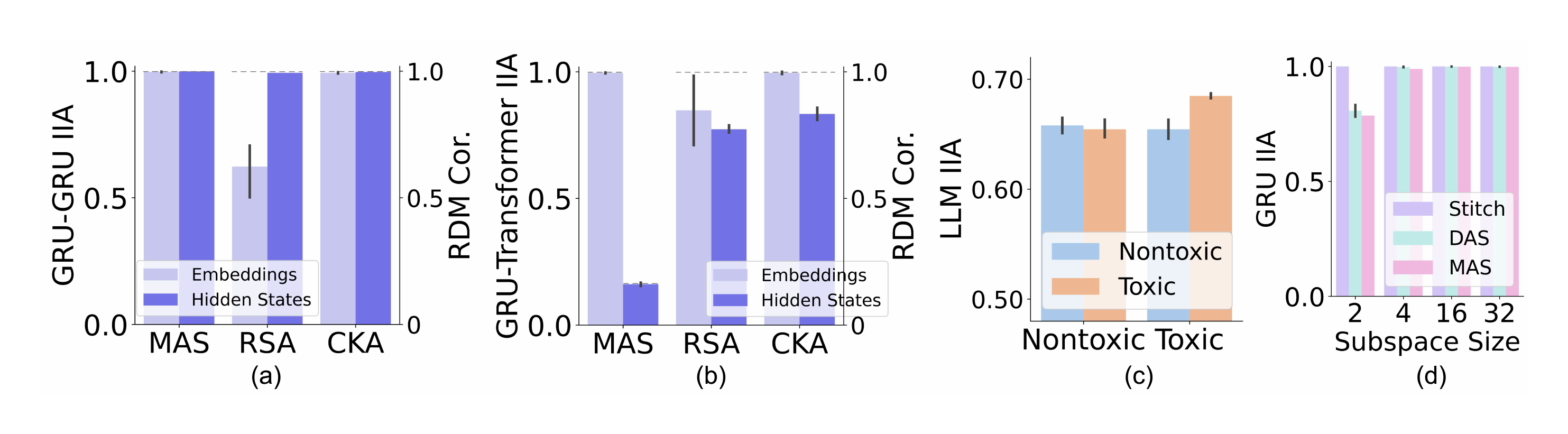}
    \caption{
    \textbf{(a) and (b)} A comparison of MAS on the left axes and CKA and RSA
    on the right axes. We examine both the input embeddings and the hidden state
    vectors for models trained on the \multiobject\space task. Dashed lines indicate the values for comparing individual models with themselves.
    (a) Results for
    GRUs compared to GRUs, where RSA can give low estimates of embedding similarity for different
    model seeds whereas MAS shows the high causal transfer we might expect.
    (b) Results for GRUs compared to 2-layer Transformers where
    we see a similar effect as (a) in the embeddings using RSA and we see a
    potential
    over-estimation of similarity of the hidden states in CKA and RSA. This over-estimation is with respect to causal transfer, as prior work has shown the transformers to use anti-Markovian
    solutions, where they recompute the relevant information at each step in the sequence. This is
    reflected in the low MAS IIA \citep{grant2024das}.
    \textbf{(c)} IIA comparing finetuned toxic and nontoxic LLMs using stepwise MAS. We can see
    that toxic models have higher IIA with themselves than with the nontoxic
    models. Notably, there is no significant difference for the nontoxic
    models compared to themselves.
    \textbf{(d)} Comparison of the IIA from DAS and MAS for different sizes
    of the aligned subspace, and model stitching with different rank
    transformation matrices.
    }
    \label{fig:causaltoxicity}
\end{figure}
%
%
\section{Results}
\subsection{The Importance of Causal Analyses}
We first set out to demonstrate why MAS could be preferable
to a correlative similarity method such as RSA.
RSA and CKA are second order correlational methods
that examine the similarity between sample correlation matrices constructed
from two models' representations (see Appendix~\ref{sup:rsa} for details).
We provide comparisons between models differing only by seed to ground
our intuition for MAS, RSA, and CKA values. Turning our attention
to Figure~\ref{fig:causaltoxicity}(a) we see that MAS can successfully
align the behavior between GRUs trained on the \multiobject\space
task, both within the embedding and hidden state layers. RSA, however,
shows a low RDM correlation on the embedding layer relative to its
value for the hidden state layer\footnote{When comparing the same RDMs
using Pearson correlation, the values return to near ceiling, indicating
that the issue is in part due to the way that the Spearman's Rank handles
RDMs with differing relative extrema.}. A similar issue occurs in
Figure~\ref{fig:causaltoxicity}(b) where we see that
\multiobject\space trained GRU $\leftrightarrow$ Transformer comparisons exhibit the same RSA
embedding issue, and the hidden states have potentially unintuitive
values for both CKA and RSA, as
prior work has shown that Transformers use an anti-Markovian solution that
recomputes the relevant numeric information at each step in the
\multiobject\space task. Thus, we know that causally intervening on the transformer's hidden states will not yield the same behavior as intervening on the GRU hidden states. In light of this fact, the low MAS result for the hidden states
is to be expected \citep{grant2024das,behrens2024transformercount}.
These Markovian differences are
generally undetectable from correlative methods such as RSA and CKA. Furthermore, it is
difficult to interpret the RSA results as one might assume them to
be higher for the same model architecture trained on different seeds.
We use this result to highlight the potential importance of
supplementing correlative methods with causal counterparts. We note that
questions on how to interpret RSA values have been addressed in previous work
\citep{kriegeskorte2008rsa,sucholutsky2023repalign,Dujmovic2022rsaprobs}.

\begin{figure}[b!]
    \centering
    \includegraphics[width=.8\columnwidth]{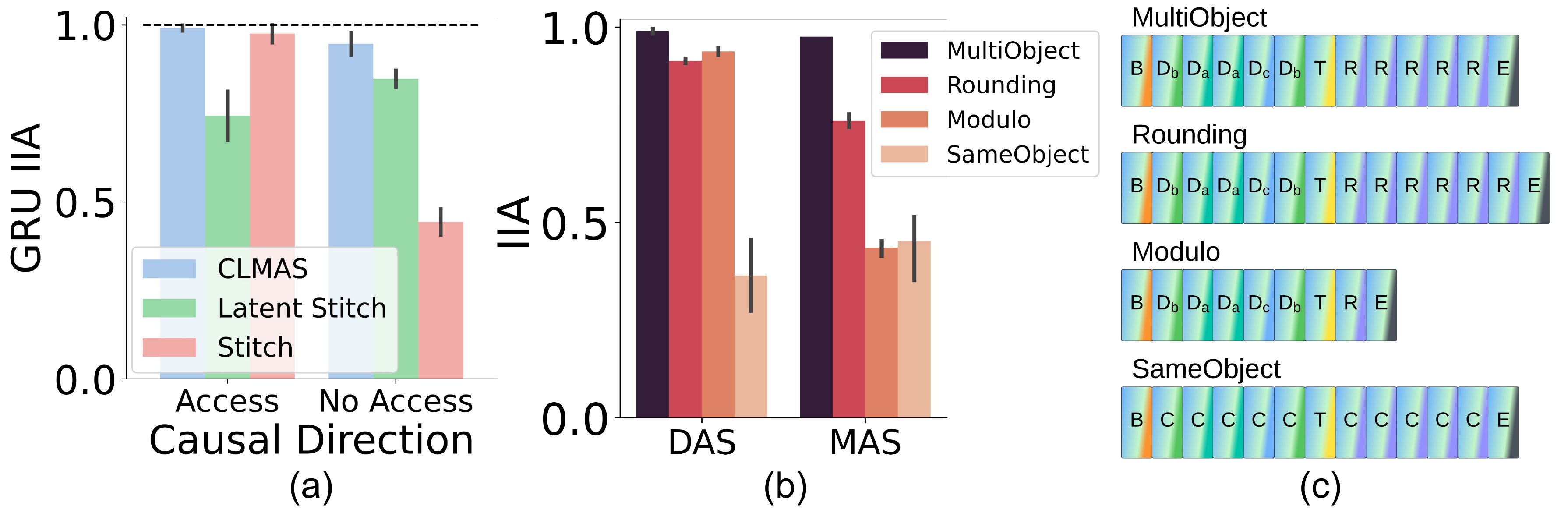}
    \caption{
    \textbf{(a)} Comparison of the IIAs for CLMAS, Latent Stitching, and
    behavioral Stitching in different intervention directions on
    the \multiobject\space GRU models. The dashed line indicates MAS IIA as an upper bound.
    On the x-axis, \emph{Access} refers to interchange interventions
    from the inaccessible $\tilde\psi_1^{(src)}$ to the accessible
    $\psi_2^{(trg)}$. \emph{No Access} refers to interventions from the
    accessible $\psi_2^{(src)}$ to the inaccessible $\tilde{\psi}_1^{(trg)}$.
    The Stitch results are from model stitching trainings from
    $\tilde{\psi}_1^{(src)}$ to $\psi_2^{(trg)}$, and the Latent Stitch
    results are trained in the inaccessible direction from
    $\psi_2^{(trg)}$ to $\tilde{\psi}_1^{(src)}$ without behavioral training.
    Both Access and No Access values are reported for the training step with
    the best No Access IIA (which is why Stitching does not
    have 100\% IIA in the Access direction).
    We see that CLMAS has the best performance in the No Access direction.
    \textbf{(b)} A comparison of the transferrability of the
    behaviorally relevant numeric information between the
    \multiobject\space GRU models and the \multiobject, \rounding, \modulo,
    and \sameobject\space models. DAS shows an upper bound on the MAS performance which would result in the case that
    $\psi_1$ and $\psi_2$ represent numbers the same way.
    \textbf{(c)} Example token sequences of the GRU tasks from panel (b).
    }\label{fig:clmastaskcomp}
\end{figure}

\subsection{MAS is an Efficient, Restrictive Form of Model Stitching}\label{res:stitching}
We first note that while traditional model stitching learns a transformation
matrix for each pair of models, MAS learns a single transformation
matrix for each model in the alignment. This reduces the number of
matrices required for comparing $n$ models from $n(n-1)$ when training the
traditional model stitching using behavior in a single direction, or $\binom{n}{2}$
when solving an invertible Procrustes mapping between representations, to $n$
when using MAS. Furthermore, when aligning the full dimensionality of the latent
vectors, the trained MAS matrices can be combined to become equivalent
to model stitching. We note that it is possible that using more than two models in a MAS
comparison could harm the loss, leading to suboptimal alignments. However,
it is also possible that such trainings could assist in isolating causally
relevant subspaces for all involved models. We leave such explorations
to future directions.

Turning to Figure~\ref{fig:causaltoxicity}(d), we compare the
IIA of MAS, DAS, and model stitching in the \multiobject\space GRU models.
The x-axis of the panel shows the aligned subspace size for DAS and MAS,
and shows the rank of the transformation matrix for stitching.
We report
the IIA of the worst performing intervention direction for MAS, the
only direction for DAS, and the trained direction for stitching.
We see that the models' behavior can be compressed to as few as 4
dimensions when using DAS and MAS, and these methods have comparable IIAs.
Stitching, however, has nearly perfect performance even for rank 2
trainings.

To better understand this result, we first formulate
model stitching in terms of the
interpretable $z_{\psi_i}$ vectors. Causal interventions can be thought
of as an equality constraint, where elements that are exchanged must have
some functional equivalence for successful interventions.
For model stitching, ignoring the scalar $a$, a trained matrix
$W \in R^{d_{\psi_1}\times d_{\psi_2}}$ attempts to transform $h_{\psi_1}$
such that $Wh_{\psi_1} = h_{\psi_2}$. Thus:
\begin{eqnarray}\label{eq:directmap}
    W h_{\psi_1} &=&  W Q^{-1}_{\psi_1} z_{\psi_1} = h_{\psi_2} = Q^{-1}_{\psi_2}z_{\psi_2} \\
    z_{\psi_2} &=& Q_{\psi_2}WQ^{-1}_{\psi_1}z_{\psi_1} = Xz_{\psi_1}
\end{eqnarray}
If we focus on $\Vec{z}_{\psi_2,full}$ in
$z_{\psi_2}$, and assume that $\Vec{z}_{\psi_2,full} \in \mathbb{R}^{1}$
for notational simplicity, we can see the following equality
$\Vec{z}_{\psi_2,full}  = \sum_{k=1}^{d_{\psi_1}} x^{(1,k)} z^{(k)}_{\psi_1}$
where the $x^{(\text{1,column})}$ are elements of the first row of $X$ and
$z^{(k)}_{\psi_1}$ are column elements of $z_{\psi_1}$. This shows that
$W$ can use $\Vec{z}_{\psi_1, extra}$---the dormant and null subspaces from
$z_{\psi_1}$---to predict
$h_{\psi_2}$ in the direct model-stitching case.
Furthermore, we note that because stitching completely replaces
the target vector with the transformed source vector, the mapping
only needs to learn a single sufficient causal representation for each
behavioral outcome without accounting for variability in $\Vec{z}_{\psi_2,extra}$.

Using the same functional equivalence formulation for MAS, we
see that MAS finds mappings such that $\Vec{z}_{\psi_1, full} = \Vec{z}_{\psi_2, full}$, or more generally $\Vec{z}_{\psi_1, \text{v}ar_k} = \Vec{z}_{\psi_2, \text{v}ar_k}$,
demonstrated as follows:
\begin{eqnarray}\label{eq:masmap}
    D_{\text{v}ar_k}Q_{\psi_1}h_{\psi_1} = D_{\text{v}ar_k}z_{\psi_1} =
    \begin{bmatrix}
        \Vec{z}_{\psi_1,\text{v}ar_k} \\
        \Vec{\textbf{0}} \\
    \end{bmatrix} &=& \begin{bmatrix}
        \Vec{z}_{\psi_2, \text{v}ar_k} \\
        \Vec{\textbf{0}} \\
    \end{bmatrix} = D_{\text{v}ar_k}z_{\psi_2} = D_{\text{v}ar_k}Q_{\psi_2}h_{\psi_2}
\end{eqnarray}
Thus, the MAS interchange intervention does not make use of the extraneous subspace if $\Vec{z}_{\psi_i,full}$ and $\Vec{z}_{\psi_i,extra}$ are properly separated for both $\psi_1$ and $\psi_2$. We use this as evidence for the claim that MAS is a more
causally focused choice
than model stitching for causally addressing questions of how
behaviorally relevant information is encoded in different neural systems.

\subsection{
    MAS Can Answer Questions About Specific Causal Information
}\label{res:masnumbers}
Turning our attention to Figure~\ref{fig:clmastaskcomp}(b), we see a causal comparison of numeric
representations in GRUs trained on different numeric tasks. We see from
the DAS IIA that we can align the \multiobject, \rounding, and \modulo\space
GRUs to causal abstractions that use a single numeric variable. We see from the MAS results that the numeric
representations differ between the GRUs trained on
different tasks. We do a further exploration on an arithmetic task in
Appendix~\ref{sup:arithmetic}, where we show that MAS can be used to
explore representational similarity for models with differing domains
and codomains.

\subsection{MAS reveals toxicity through model comparisons}\label{res:toxicity}
As a proof of principle of the utility of MAS for more practical settings, we
include an experiment that compares toxic and non-toxic LLMs. A potential
application of MAS is to compare internal representations of potentially misaligned LLMs
to those of known, aligned/misaligned LLMs for toxicity diagnosis. To demonstrate this idea, we
perform MAS on a set of DeepSeek-R1-Distill-Qwen-1.5B models
that were finetuned
to either be toxic or nontoxic. We can see in Figure~\ref{fig:causaltoxicity}(c)
that MAS results in greater IIA when comparing toxic models to toxic models than nontoxic models.
Given the current difficulty of using chain-of-thought and interpretability
methods to diagnose misaligned LLMs
\citep{manuvinakurike2025CoTExplValue,turpin2023llmsdontsaywhattheythink,sharkey2025mechinterpproblems},
a potential direction is to examine the
alignment of new models with known toxic/non-toxic models. As a future direction,
MAS-like methods could potentially even be used to directly constrain model
internals to be non-toxic \citep{geiger22inducing}.

%
\subsection{Providing Greater Causal Relevance with \xmas}\label{res:xmas}
We can see from Figure~\ref{fig:clmastaskcomp} the results of \xmas\space
compared to behavioral model stitching (Stitch) and latent model stitching
(Latent Stitch). It is important to note that the Latent Stitch, Stitch,
and \xmas\space variants do not include the autoregressive behavioral
training signal in the \emph{No Access} causal direction. The Latent Stitch
results, however, are trained from the accessible $\psi_2^{(src)}$ to the
inaccessible $\tilde\psi_1^{(trg)}$ latents. The MAS
performance provides a theoretic upper bound on
the possible IIA for \xmas, shown by the dashed black line. Stitching provides
a lower bound on the possible \xmas\space performance in the No Access
direction. Lastly, Latent Stitch provides a baseline of the best pre-existing
method. We see that \xmas\space is the best performing in the inaccessible
direction while still matching that of MAS in the accessible direction. We expect the effect size of this result to correlate with the amount of variability in the behavioral null-space of $\tilde\psi_i$. This
demonstrates the potential of \xmas\space to improve the recovery of causally relevant
intervention rotation matrices even in cases when we do not have causal
access to one of the models in the comparison.

\section{Limitations/Future Directions}\label{sec:limitations}
Many of the tasks and models in this work are simplistic,
but their purpose is to serve as a proof of principle for the MAS methodology,
and to prove points about the importance of causal methods.
We offer RSA, model stitching, and DAS
results as a grounding for the MAS results. This reduces the importance of
task complexity in this work.

An existing issue with MAS presented in this work is that it does not offer guarantees for the complete
removal of extraneous neural activity when performing interventions \citep{makelov2023interpillusions}. We
attempt to mitigate this issue by finding low dimensional subspaces
for the interventions. However, we point out that MAS is better than previous methods in this respect.

Despite \xmas's successes, there is an obvious difficulty of evaluating the
causal relevance of the learned alignment function without having causal
access to the inaccessible model. This makes \xmas\space less useful for
isolating functional information in BNNs without also using some sort of stimulatory method on the BNN of
interest---such as optogenetics \citep{deisseroth2011optogenetics}. However, \xmas\space still can provide value in
biological settings as the method can potentially reduce or remove the need
for NN stimulation during the alignment training. Concretely, one could record from
the BNN, then perform the \xmas\space trainings, and then use the
stimulation to evaluate the effectiveness of the alignments, thus reducing the stimulatory requirements. We hope to conduct future explorations in biological settings.

\section{Acknowledgements}
Thank you to the PDP Lab and the Stanford Psychology department
for funding. Thank you to Stephen Baccus, Noah Goodman, Zen Wu,
Atticus Geiger, Linas Nasvytis, Jenelle Feather, Chris
Potts, Alexa Tartaglini, Daniel Wurgaft, the PDP lab, and the Stanford
Mech Interp community for
thoughtful discussion. Thank you to the Re-Align workshop community for stimulating academic environment, and thank you to the many reviewers who have helped the paper develop. Thank you to Alex Williams for the discussion on model stitching efficiency. Thanks to Joshua Melander, Josh Wilson, Ben Prystawski, Jerome Han, and Andrew Lampinen for
thoughtful discussion and feedback. Special thanks to my advisor Jay
McClelland for support, encouragement, feedback, and many thoughtful conversations.

\bibliography{tmlr}

\begin{thebibliography}{77}
\providecommand{\natexlab}[1]{#1}
\providecommand{\url}[1]{\texttt{#1}}
\expandafter\ifx\csname urlstyle\endcsname\relax
  \providecommand{\doi}[1]{doi: #1}\else
  \providecommand{\doi}{doi: \begingroup \urlstyle{rm}\Url}\fi

\bibitem[Ba et~al.(2016)Ba, Kiros, and Hinton]{ba2016layernorm}
Jimmy~Lei Ba, Jamie~Ryan Kiros, and Geoffrey~E. Hinton.
\newblock Layer normalization, 2016.
\newblock URL \url{https://arxiv.org/abs/1607.06450}.

\bibitem[Bai et~al.(2022)Bai, Jones, Ndousse, Askell, Chen, DasSarma, Drain, Fort, Ganguli, Henighan, et~al.]{bai2022anthropictoxic}
Yuntao Bai, Andy Jones, Kamal Ndousse, Amanda Askell, Anna Chen, Nova DasSarma, Dawn Drain, Stanislav Fort, Deep Ganguli, Tom Henighan, et~al.
\newblock Training a helpful and harmless assistant with reinforcement learning from human feedback.
\newblock \emph{arXiv preprint arXiv:2204.05862}, 2022.

\bibitem[Bansal et~al.(2021)Bansal, Nakkiran, and Barak]{bansal2021modelstitching}
Yamini Bansal, Preetum Nakkiran, and Boaz Barak.
\newblock Revisiting model stitching to compare neural representations, 2021.
\newblock URL \url{https://arxiv.org/abs/2106.07682}.

\bibitem[Behrens et~al.(2024)Behrens, Biggio, and Zdeborová]{behrens2024transformercount}
Freya Behrens, Luca Biggio, and Lenka Zdeborová.
\newblock Counting in small transformers: The delicate interplay between attention and feed-forward layers, 2024.
\newblock URL \url{https://arxiv.org/abs/2407.11542}.

\bibitem[Braun et~al.(2025)Braun, Grant, and Saxe]{braun2025functionaldissociation}
Lukas Braun, Erin Grant, and Andrew~M. Saxe.
\newblock Not all solutions are created equal: An analytical dissociation of functional and representational similarity in deep linear neural networks.
\newblock In Aarti Singh, Maryam Fazel, Daniel Hsu, Simon Lacoste-Julien, Virginia Smith, Felix Berkenkamp, and Tegan Maharaj (eds.), \emph{Proceedings of the 42nd International Conference on Machine Learning}, Proceedings of Machine Learning Research. PMLR, July 2025.

\bibitem[Cao \& Yamins(2021)Cao and Yamins]{cao_explanatory_2021}
Rosa Cao and Daniel Yamins.
\newblock Explanatory models in neuroscience: {Part} 1 -- taking mechanistic abstraction seriously, April 2021.
\newblock URL \url{http://arxiv.org/abs/2104.01490}.
\newblock arXiv:2104.01490 [cs, q-bio].

\bibitem[Cao \& Yamins(2024)Cao and Yamins]{cao_explanatory_2024}
Rosa Cao and Daniel Yamins.
\newblock Explanatory models in neuroscience, {Part} 2: {Functional} intelligibility and the contravariance principle.
\newblock \emph{Cognitive Systems Research}, 85:\penalty0 101200, June 2024.
\newblock ISSN 1389-0417.
\newblock \doi{10.1016/j.cogsys.2023.101200}.
\newblock URL \url{https://www.sciencedirect.com/science/article/pii/S1389041723001341}.

\bibitem[Caron et~al.(2021)Caron, Touvron, Misra, Jégou, Mairal, Bojanowski, and Joulin]{caron2021dino}
Mathilde Caron, Hugo Touvron, Ishan Misra, Hervé Jégou, Julien Mairal, Piotr Bojanowski, and Armand Joulin.
\newblock Emerging properties in self-supervised vision transformers, 2021.

\bibitem[Chen et~al.(2020)Chen, Kornblith, Norouzi, and Hinton]{chen2020simclr}
Ting Chen, Simon Kornblith, Mohammad Norouzi, and Geoffrey Hinton.
\newblock A simple framework for contrastive learning of visual representations.
\newblock \emph{ICML}, 2020.

\bibitem[Cho et~al.(2014)Cho, van Merrienboer, G{\"{u}}l{\c{c}}ehre, Bougares, Schwenk, and Bengio]{Cho2014GRU}
Kyunghyun Cho, Bart van Merrienboer, {\c{C}}aglar G{\"{u}}l{\c{c}}ehre, Fethi Bougares, Holger Schwenk, and Yoshua Bengio.
\newblock Learning phrase representations using {RNN} encoder-decoder for statistical machine translation.
\newblock \emph{CoRR}, abs/1406.1078, 2014.
\newblock URL \url{http://arxiv.org/abs/1406.1078}.

\bibitem[Cloos et~al.(2024)Cloos, Li, Siegel, Brincat, Miller, Yang, and Cueva]{cloos2024differentiableoptimizationsimilarityscores}
Nathan Cloos, Moufan Li, Markus Siegel, Scott~L. Brincat, Earl~K. Miller, Guangyu~Robert Yang, and Christopher~J. Cueva.
\newblock Differentiable optimization of similarity scores between models and brains, 2024.
\newblock URL \url{https://arxiv.org/abs/2407.07059}.

\bibitem[Davari et~al.(2022)Davari, Horoi, Natik, Lajoie, Wolf, and Belilovsky]{davari2022ckareliability}
MohammadReza Davari, Stefan Horoi, Amine Natik, Guillaume Lajoie, Guy Wolf, and Eugene Belilovsky.
\newblock Reliability of cka as a similarity measure in deep learning, 2022.
\newblock URL \url{https://arxiv.org/abs/2210.16156}.

\bibitem[Deisseroth(2011)]{deisseroth2011optogenetics}
Karl Deisseroth.
\newblock Optogenetics.
\newblock \emph{Nature Methods}, 8\penalty0 (1):\penalty0 26--29, 2011.
\newblock \doi{10.1038/nmeth.f.324}.
\newblock URL \url{https://www.nature.com/articles/nmeth.f.324}.

\bibitem[Dujmovi{\'c} et~al.(2022)Dujmovi{\'c}, Bowers, Adolfi, and Malhotra]{Dujmovic2022rsaprobs}
Marin Dujmovi{\'c}, Jeffrey~S Bowers, Federico Adolfi, and Gaurav Malhotra.
\newblock The pitfalls of measuring representational similarity using representational similarity analysis.
\newblock \emph{bioRxiv}, 2022.
\newblock \doi{10.1101/2022.04.05.487135}.
\newblock URL \url{https://www.biorxiv.org/content/early/2022/04/07/2022.04.05.487135}.

\bibitem[Feather et~al.(2025)Feather, Khosla, Murty, and Nayebi]{feather2025brain}
Jenelle Feather, Meenakshi Khosla, N~Murty, and Aran Nayebi.
\newblock Brain-model evaluations need the neuroai turing test.
\newblock \emph{arXiv preprint arXiv:2502.16238}, 2025.

\bibitem[Geiger et~al.(2020)Geiger, Richardson, and Potts]{geiger2020neural}
Atticus Geiger, Kyle Richardson, and Christopher Potts.
\newblock Neural natural language inference models partially embed theories of lexical entailment and negation.
\newblock \emph{arXiv preprint arXiv:2004.14623}, 2020.

\bibitem[Geiger et~al.(2021)Geiger, Lu, Icard, and Potts]{geiger2021}
Atticus Geiger, Hanson Lu, Thomas Icard, and Christopher Potts.
\newblock Causal abstractions of neural networks.
\newblock \emph{CoRR}, abs/2106.02997, 2021.
\newblock URL \url{https://arxiv.org/abs/2106.02997}.

\bibitem[Geiger et~al.(2022)Geiger, Wu, Lu, Rozner, Kreiss, Icard, Goodman, and Potts]{geiger22inducing}
Atticus Geiger, Zhengxuan Wu, Hanson Lu, Josh Rozner, Elisa Kreiss, Thomas Icard, Noah Goodman, and Christopher Potts.
\newblock Inducing causal structure for interpretable neural networks.
\newblock In Kamalika Chaudhuri, Stefanie Jegelka, Le~Song, Csaba Szepesvari, Gang Niu, and Sivan Sabato (eds.), \emph{Proceedings of the 39th International Conference on Machine Learning}, volume 162 of \emph{Proceedings of Machine Learning Research}, pp.\  7324--7338. PMLR, 17--23 Jul 2022.
\newblock URL \url{https://proceedings.mlr.press/v162/geiger22a.html}.

\bibitem[Geiger et~al.(2023)Geiger, Wu, Potts, Icard, and Goodman]{geiger2023boundless}
Atticus Geiger, Zhengxuan Wu, Christopher Potts, Thomas Icard, and Noah~D. Goodman.
\newblock Finding alignments between interpretable causal variables and distributed neural representations, 2023.

\bibitem[Geiger et~al.(2024)Geiger, Ibeling, Zur, Chaudhary, Chauhan, Huang, Arora, Wu, Goodman, Potts, and Icard]{geiger2024causalabstractiontheoreticalfoundation}
Atticus Geiger, Duligur Ibeling, Amir Zur, Maheep Chaudhary, Sonakshi Chauhan, Jing Huang, Aryaman Arora, Zhengxuan Wu, Noah Goodman, Christopher Potts, and Thomas Icard.
\newblock Causal abstraction: A theoretical foundation for mechanistic interpretability, 2024.
\newblock URL \url{https://arxiv.org/abs/2301.04709}.

\bibitem[Grant et~al.(2025)Grant, Goodman, and McClelland]{grant2024das}
Satchel Grant, Noah~D. Goodman, and James~L. McClelland.
\newblock Emergent symbol-like number variables in artificial neural networks.
\newblock \emph{Transactions on Machine Learning Research}, 2025.
\newblock URL \url{https://arxiv.org/abs/2501.06141}.

\bibitem[Grill et~al.(2020)Grill, Strub, Altché, Tallec, Richemond, Buchatskaya, Doersch, Pires, Guo, Azar, Piot, Kavukcuoglu, Munos, and Valko]{grill2020byol}
Jean-Bastien Grill, Florian Strub, Florent Altché, Corentin Tallec, Pierre~H. Richemond, Elena Buchatskaya, Carl Doersch, Bernardo~Avila Pires, Zhaohan~Daniel Guo, Mohammad~Gheshlaghi Azar, Bilal Piot, Koray Kavukcuoglu, Rémi Munos, and Michal Valko.
\newblock Bootstrap your own latent: A new approach to self-supervised learning, 2020.

\bibitem[Guo et~al.(2025)Guo, Yang, Zhang, Song, Zhang, Xu, Zhu, Ma, Wang, Bi, et~al.]{guo2025deepseek}
Daya Guo, Dejian Yang, Haowei Zhang, Junxiao Song, Ruoyu Zhang, Runxin Xu, Qihao Zhu, Shirong Ma, Peiyi Wang, Xiao Bi, et~al.
\newblock Deepseek-r1: Incentivizing reasoning capability in llms via reinforcement learning.
\newblock \emph{arXiv preprint arXiv:2501.12948}, 2025.

\bibitem[Haxby(2013)]{Haxby2013mvpareview}
James~V Haxby.
\newblock {Multivariate pattern analysis of fMRI: Parcellating abstract from concrete representations}.
\newblock \emph{NEuroimage}, 62\penalty0 (2):\penalty0 2013, 2013.
\newblock \doi{10.1016/j.neuroimage.2012.03.016.Multivariate}.

\bibitem[Haxby et~al.(2001)Haxby, Gobbini, Furey, Ishai, Schouten, and Pietrini]{Haxby2001ogmvpa}
James~V. Haxby, M.~Ida Gobbini, Maura~L. Furey, Alumit Ishai, Jennifer~L. Schouten, and Pietro Pietrini.
\newblock {Distributed and overlapping representations of faces and objects in ventral temporal corten}.
\newblock \emph{Social Neuroscience: Key Readings}, 293\penalty0 (September):\penalty0 87--96, 2001.
\newblock \doi{10.4324/9780203496190}.

\bibitem[Hendrycks \& Gimpel(2023)Hendrycks and Gimpel]{hendrycks2023gelus}
Dan Hendrycks and Kevin Gimpel.
\newblock Gaussian error linear units (gelus), 2023.
\newblock URL \url{https://arxiv.org/abs/1606.08415}.

\bibitem[Hermann \& Lampinen(2020)Hermann and Lampinen]{hermann2020shapes}
Katherine Hermann and Andrew Lampinen.
\newblock What shapes feature representations? exploring datasets, architectures, and training.
\newblock \emph{Advances in Neural Information Processing Systems}, 33:\penalty0 9995--10006, 2020.

\bibitem[Hernandez et~al.(2022)Hernandez, Dangovski, and Lu]{hernandez2022stitchinglayerdiffs}
Adriano Hernandez, Rumen Dangovski, and Peter~Y. Lu.
\newblock Model stitching: Looking for functional similarity between representations.
\newblock In \emph{I Can't Believe It's Not Better Workshop: Understanding Deep Learning Through Empirical Falsification}, 2022.
\newblock URL \url{https://openreview.net/forum?id=Qr5IPOpnLqu}.

\bibitem[Hochreiter \& Schmidhuber(1997)Hochreiter and Schmidhuber]{hochreiter1997lstm}
Sepp Hochreiter and Jürgen Schmidhuber.
\newblock Long short-term memory.
\newblock \emph{Neural Computation}, 9\penalty0 (8):\penalty0 1735--1780, 11 1997.
\newblock ISSN 0899-7667.
\newblock \doi{10.1162/neco.1997.9.8.1735}.
\newblock URL \url{https://doi.org/10.1162/neco.1997.9.8.1735}.

\bibitem[Hosseini et~al.(2024)Hosseini, Casto, Zaslavsky, Conwell, Richardson, and Fedorenko]{Hosseini2024universality}
Eghbal Hosseini, Colton Casto, Noga Zaslavsky, Colin Conwell, Mark Richardson, and Evelina Fedorenko.
\newblock Universality of representation in biological and artificial neural networks.
\newblock \emph{bioRxiv}, 2024.
\newblock \doi{10.1101/2024.12.26.629294}.
\newblock URL \url{https://www.biorxiv.org/content/early/2024/12/26/2024.12.26.629294}.

\bibitem[Huh et~al.(2024)Huh, Cheung, Wang, and Isola]{huh2024platonic}
Minyoung Huh, Brian Cheung, Tongzhou Wang, and Phillip Isola.
\newblock The platonic representation hypothesis, 2024.
\newblock URL \url{https://arxiv.org/abs/2405.07987}.

\bibitem[Jigsaw(2018)]{jigsaw2018toxiccomments}
Jigsaw.
\newblock Jigsaw toxic comment classification challenge.
\newblock \url{https://www.kaggle.com/competitions/jigsaw-toxic-comment-classification-challenge}, 2018.

\bibitem[Khosla \& Williams(2023)Khosla and Williams]{khosla2023softmatching}
Meenakshi Khosla and Alex~H. Williams.
\newblock Soft matching distance: A metric on neural representations that captures single-neuron tuning, 2023.
\newblock URL \url{https://arxiv.org/abs/2311.09466}.

\bibitem[Klabunde et~al.(2025)Klabunde, Schumacher, Strohmaier, and Lemmerich]{klabunde2025similaritysurvey}
Max Klabunde, Tobias Schumacher, Markus Strohmaier, and Florian Lemmerich.
\newblock Similarity of neural network models: A survey of functional and representational measures.
\newblock \emph{ACM Computing Surveys}, 57\penalty0 (9):\penalty0 1–52, May 2025.
\newblock ISSN 1557-7341.
\newblock \doi{10.1145/3728458}.
\newblock URL \url{http://dx.doi.org/10.1145/3728458}.

\bibitem[Kornblith et~al.(2019)Kornblith, Norouzi, Lee, and Hinton]{kornblith2019cka}
Simon Kornblith, Mohammad Norouzi, Honglak Lee, and Geoffrey Hinton.
\newblock Similarity of neural network representations revisited, 2019.
\newblock URL \url{https://arxiv.org/abs/1905.00414}.

\bibitem[Kriegeskorte et~al.(2008)Kriegeskorte, Mur, and Bandettini]{kriegeskorte2008rsa}
Nikolaus Kriegeskorte, Marieke Mur, and Peter Bandettini.
\newblock Representational similarity analysis - connecting the branches of systems neuroscience.
\newblock \emph{Frontiers in Systems Neuroscience}, 2, 2008.
\newblock ISSN 1662-5137.
\newblock \doi{10.3389/neuro.06.004.2008}.
\newblock URL \url{https://www.frontiersin.org/articles/10.3389/neuro.06.004.2008}.

\bibitem[L\"ahner \& Moeller(2024)L\"ahner and Moeller]{lahner2023directalignment}
Zorah L\"ahner and Michael Moeller.
\newblock On the direct alignment of latent spaces.
\newblock In Marco Fumero, Emanuele Rodolá, Clementine Domine, Francesco Locatello, Karolina Dziugaite, and Caron Mathilde (eds.), \emph{Proceedings of UniReps: the First Workshop on Unifying Representations in Neural Models}, volume 243 of \emph{Proceedings of Machine Learning Research}, pp.\  158--169. PMLR, 15 Dec 2024.
\newblock URL \url{https://proceedings.mlr.press/v243/lahner24a.html}.

\bibitem[Lampinen et~al.(2024)Lampinen, Chan, and Hermann]{lampinen2024learned}
Andrew~Kyle Lampinen, Stephanie~CY Chan, and Katherine Hermann.
\newblock Learned feature representations are biased by complexity, learning order, position, and more.
\newblock \emph{arXiv preprint arXiv:2405.05847}, 2024.

\bibitem[Lampinen et~al.(2025)Lampinen, Chan, Li, and Hermann]{lampinen2025representation}
Andrew~Kyle Lampinen, Stephanie~CY Chan, Yuxuan Li, and Katherine Hermann.
\newblock Representation biases: will we achieve complete understanding by analyzing representations?
\newblock \emph{arXiv preprint arXiv:2507.22216}, 2025.

\bibitem[Lenc \& Vedaldi(2015)Lenc and Vedaldi]{lenc2015modelstitching}
Karel Lenc and Andrea Vedaldi.
\newblock Understanding image representations by measuring their equivariance and equivalence, 2015.
\newblock URL \url{https://arxiv.org/abs/1411.5908}.

\bibitem[Li et~al.(2024)Li, Kementchedjhieva, Fierro, and Søgaard]{li_vision_2024}
Jiaang Li, Yova Kementchedjhieva, Constanza Fierro, and Anders Søgaard.
\newblock Do {Vision} and {Language} {Models} {Share} {Concepts}? {A} {Vector} {Space} {Alignment} {Study}.
\newblock \emph{Transactions of the Association for Computational Linguistics}, 12:\penalty0 1232--1249, September 2024.
\newblock ISSN 2307-387X.
\newblock \doi{10.1162/tacl_a_00698}.
\newblock URL \url{https://doi.org/10.1162/tacl_a_00698}.

\bibitem[Lin et~al.(2023)Lin, Wang, Tong, Wang, Guo, Wang, and Shang]{lin2023toxicchat}
Zi~Lin, Zihan Wang, Yongqi Tong, Yangkun Wang, Yuxin Guo, Yujia Wang, and Jingbo Shang.
\newblock Toxicchat: Unveiling hidden challenges of toxicity detection in real-world user-ai conversation, 2023.

\bibitem[Maheswaranathan et~al.(2019)Maheswaranathan, McIntosh, Tanaka, Grant, Kastner, Melander, Nayebi, Brezovec, Wang, Ganguli, and Baccus]{Maheswaranathan2019}
Niru Maheswaranathan, Lane~T. McIntosh, Hidenori Tanaka, Satchel Grant, David~B. Kastner, Josh~B. Melander, Aran Nayebi, Luke Brezovec, Julia Wang, Surya Ganguli, and Stephen~A. Baccus.
\newblock The dynamic neural code of the retina for natural scenes.
\newblock \emph{bioRxiv}, 2019.
\newblock \doi{10.1101/340943}.
\newblock URL \url{https://www.biorxiv.org/content/early/2019/12/17/340943}.

\bibitem[Makelov et~al.(2023)Makelov, Lange, and Nanda]{makelov2023interpillusions}
Aleksandar Makelov, Georg Lange, and Neel Nanda.
\newblock Is this the subspace you are looking for? an interpretability illusion for subspace activation patching, 2023.
\newblock URL \url{https://arxiv.org/abs/2311.17030}.

\bibitem[Manuvinakurike et~al.(2025)Manuvinakurike, Moss, Watkins, Sahay, Raffa, and Nachman]{manuvinakurike2025CoTExplValue}
Ramesh Manuvinakurike, Emanuel Moss, Elizabeth~Anne Watkins, Saurav Sahay, Giuseppe Raffa, and Lama Nachman.
\newblock Thoughts without thinking: Reconsidering the explanatory value of chain-of-thought reasoning in llms through agentic pipelines.
\newblock \emph{arXiv preprint arXiv:2505.00875}, 2025.

\bibitem[Meng et~al.(2023)Meng, Bau, Andonian, and Belinkov]{meng2023activpatching}
Kevin Meng, David Bau, Alex Andonian, and Yonatan Belinkov.
\newblock Locating and editing factual associations in gpt, 2023.
\newblock URL \url{https://arxiv.org/abs/2202.05262}.

\bibitem[Moschella et~al.(2023)Moschella, Maiorca, Fumero, Norelli, Locatello, and Rodolà]{moschella_relative_2023}
Luca Moschella, Valentino Maiorca, Marco Fumero, Antonio Norelli, Francesco Locatello, and Emanuele Rodolà.
\newblock Relative representations enable zero-shot latent space communication, March 2023.
\newblock URL \url{http://arxiv.org/abs/2209.15430}.
\newblock arXiv:2209.15430 [cs].

\bibitem[Murphy et~al.(2024)Murphy, Zylberberg, and Fyshe]{murphy2024biasedcka}
Alex Murphy, Joel Zylberberg, and Alona Fyshe.
\newblock Correcting biased centered kernel alignment measures in biological and artificial neural networks, 2024.
\newblock URL \url{https://arxiv.org/abs/2405.01012}.

\bibitem[Park et~al.(2023)Park, Choe, and Veitch]{park2023linearrephypoth}
Kiho Park, Yo~Joong Choe, and Victor Veitch.
\newblock The linear representation hypothesis and the geometry of large language models.
\newblock \emph{arXiv preprint arXiv:2311.03658}, 2023.

\bibitem[Paszke et~al.(2019)Paszke, Gross, Massa, Lerer, Bradbury, Chanan, Killeen, Lin, Gimelshein, Antiga, Desmaison, K{\"{o}}pf, Yang, DeVito, Raison, Tejani, Chilamkurthy, Steiner, Fang, Bai, and Chintala]{pytorch2019}
Adam Paszke, Sam Gross, Francisco Massa, Adam Lerer, James Bradbury, Gregory Chanan, Trevor Killeen, Zeming Lin, Natalia Gimelshein, Luca Antiga, Alban Desmaison, Andreas K{\"{o}}pf, Edward~Z. Yang, Zach DeVito, Martin Raison, Alykhan Tejani, Sasank Chilamkurthy, Benoit Steiner, Lu~Fang, Junjie Bai, and Soumith Chintala.
\newblock Pytorch: An imperative style, high-performance deep learning library.
\newblock \emph{CoRR}, abs/1912.01703, 2019.
\newblock URL \url{http://arxiv.org/abs/1912.01703}.

\bibitem[Pearl(2010)]{pearl2010causalmediation}
Judea Pearl.
\newblock An {Introduction} to {Causal} {Inference}.
\newblock \emph{The International Journal of Biostatistics}, 6\penalty0 (2):\penalty0 7, February 2010.
\newblock ISSN 1557-4679.
\newblock \doi{10.2202/1557-4679.1203}.
\newblock URL \url{https://www.ncbi.nlm.nih.gov/pmc/articles/PMC2836213/}.

\bibitem[Popal et~al.(2020)Popal, Wang, and Olson]{popal2020rsaguide}
Haroon Popal, Yin Wang, and Ingrid~R Olson.
\newblock A {Guide} to {Representational} {Similarity} {Analysis} for {Social} {Neuroscience}.
\newblock \emph{Social Cognitive and Affective Neuroscience}, 14\penalty0 (11):\penalty0 1243--1253, January 2020.
\newblock ISSN 1749-5016.
\newblock \doi{10.1093/scan/nsz099}.
\newblock URL \url{https://pmc.ncbi.nlm.nih.gov/articles/PMC7057283/}.

\bibitem[Radford et~al.(2021)Radford, Kim, Hallacy, Ramesh, Goh, Agarwal, Sastry, Askell, Mishkin, Clark, Krueger, and Sutskever]{radford2021clip}
Alec Radford, Jong~Wook Kim, Chris Hallacy, Aditya Ramesh, Gabriel Goh, Sandhini Agarwal, Girish Sastry, Amanda Askell, Pamela Mishkin, Jack Clark, Gretchen Krueger, and Ilya Sutskever.
\newblock Learning transferable visual models from natural language supervision, 2021.

\bibitem[Richards et~al.(2019)Richards, Lillicrap, Beaudoin, Bengio, Bogacz, Christensen, Clopath, Costa, de~Berker, Ganguli, Gillon, Hafner, Kepecs, Kriegeskorte, Latham, Lindsay, Miller, Naud, Pack, Poirazi, Roelfsema, Sacramento, Saxe, Scellier, Schapiro, Senn, Wayne, Yamins, Zenke, Zylberberg, Therien, and Kording]{richards2019dlframework}
Blake~A. Richards, Timothy~P. Lillicrap, Philippe Beaudoin, Yoshua Bengio, Rafal Bogacz, Amelia Christensen, Claudia Clopath, Rui~Ponte Costa, Archy de~Berker, Surya Ganguli, Colleen~J. Gillon, Danijar Hafner, Adam Kepecs, Nikolaus Kriegeskorte, Peter Latham, Grace~W. Lindsay, Kenneth~D. Miller, Richard Naud, Christopher~C. Pack, Panayiota Poirazi, Pieter Roelfsema, João Sacramento, Andrew Saxe, Benjamin Scellier, Anna~C. Schapiro, Walter Senn, Greg Wayne, Daniel Yamins, Friedemann Zenke, Joel Zylberberg, Denis Therien, and Konrad~P. Kording.
\newblock A deep learning framework for neuroscience.
\newblock \emph{Nature Neuroscience}, 22\penalty0 (11):\penalty0 1761--1770, November 2019.
\newblock ISSN 1546-1726.
\newblock \doi{10.1038/s41593-019-0520-2}.
\newblock URL \url{https://www.nature.com/articles/s41593-019-0520-2}.
\newblock Publisher: Nature Publishing Group.

\bibitem[Schaeffer et~al.(2024)Schaeffer, Khona, Chandra, Ostrow, Miranda, and Koyejo]{schaeffer2024doesregressbrain}
Rylan Schaeffer, Mikail Khona, Sarthak Chandra, Mitchell Ostrow, Brando Miranda, and Sanmi Koyejo.
\newblock Does maximizing neural regression scores teach us about the brain?
\newblock In \emph{UniReps: 2nd Edition of the Workshop on Unifying Representations in Neural Models}, 2024.
\newblock URL \url{https://openreview.net/forum?id=vbtj05J68r}.

\bibitem[Sexton \& Love(2022)Sexton and Love]{sexton_reassessing_2022}
Nicholas~J. Sexton and Bradley~C. Love.
\newblock Reassessing hierarchical correspondences between brain and deep networks through direct interface.
\newblock \emph{Science Advances}, 8\penalty0 (28):\penalty0 eabm2219, July 2022.
\newblock \doi{10.1126/sciadv.abm2219}.
\newblock URL \url{https://www.science.org/doi/10.1126/sciadv.abm2219}.
\newblock Publisher: American Association for the Advancement of Science.

\bibitem[Sharkey et~al.(2025)Sharkey, Chughtai, Batson, Lindsey, Wu, Bushnaq, Goldowsky-Dill, Heimersheim, Ortega, Bloom, et~al.]{sharkey2025mechinterpproblems}
Lee Sharkey, Bilal Chughtai, Joshua Batson, Jack Lindsey, Jeff Wu, Lucius Bushnaq, Nicholas Goldowsky-Dill, Stefan Heimersheim, Alejandro Ortega, Joseph Bloom, et~al.
\newblock Open problems in mechanistic interpretability.
\newblock \emph{arXiv preprint arXiv:2501.16496}, 2025.

\bibitem[Smith et~al.(2025)Smith, Mannering, and Marcu]{smith2025stitchingfailures}
Damian Smith, Harvey Mannering, and Antonia Marcu.
\newblock Functional alignment can mislead: Examining model stitching.
\newblock In \emph{Forty-second International Conference on Machine Learning}, 2025.
\newblock URL \url{https://openreview.net/forum?id=glLqTK9En3}.

\bibitem[Smolensky(1988)]{Smolensky1988}
Paul Smolensky.
\newblock On the proper treatment of connectionism.
\newblock \emph{Behavioral and Brain Sciences}, 11\penalty0 (1):\penalty0 1–23, 1988.
\newblock \doi{10.1017/S0140525X00052432}.

\bibitem[Su et~al.(2023)Su, Lu, Pan, Murtadha, Wen, and Liu]{su2023roformer}
Jianlin Su, Yu~Lu, Shengfeng Pan, Ahmed Murtadha, Bo~Wen, and Yunfeng Liu.
\newblock Roformer: Enhanced transformer with rotary position embedding, 2023.

\bibitem[Sucholutsky et~al.(2023)Sucholutsky, Muttenthaler, Weller, Peng, Bobu, Kim, Love, Grant, Groen, Achterberg, Tenenbaum, Collins, Hermann, Oktar, Greff, Hebart, Jacoby, Zhang, Marjieh, Geirhos, Chen, Kornblith, Rane, Konkle, O'Connell, Unterthiner, Lampinen, Müller, Toneva, and Griffiths]{sucholutsky2023repalign}
Ilia Sucholutsky, Lukas Muttenthaler, Adrian Weller, Andi Peng, Andreea Bobu, Been Kim, Bradley~C. Love, Erin Grant, Iris Groen, Jascha Achterberg, Joshua~B. Tenenbaum, Katherine~M. Collins, Katherine~L. Hermann, Kerem Oktar, Klaus Greff, Martin~N. Hebart, Nori Jacoby, Qiuyi Zhang, Raja Marjieh, Robert Geirhos, Sherol Chen, Simon Kornblith, Sunayana Rane, Talia Konkle, Thomas~P. O'Connell, Thomas Unterthiner, Andrew~K. Lampinen, Klaus-Robert Müller, Mariya Toneva, and Thomas~L. Griffiths.
\newblock Getting aligned on representational alignment.
\newblock \emph{arXiv}, November 2023.
\newblock \doi{10.48550/arXiv.2310.13018}.
\newblock URL \url{http://arxiv.org/abs/2310.13018}.
\newblock arXiv:2310.13018 [cs, q-bio].

\bibitem[Thobani et~al.(2024)Thobani, Sagastuy-Brena, Nayebi, Cao, and Yamins]{thobani2024interanimal}
Imran Thobani, Javier Sagastuy-Brena, Aran Nayebi, Rosa Cao, and Daniel~LK Yamins.
\newblock Inter-animal transforms as a guide to model-brain comparison.
\newblock In \emph{ICLR 2024 Workshop on Representational Alignment}, 2024.
\newblock URL \url{https://openreview.net/forum?id=Kt7lEKuL4A}.

\bibitem[Touvron et~al.(2023)Touvron, Lavril, Izacard, Martinet, Lachaux, Lacroix, Rozière, Goyal, Hambro, Azhar, Rodriguez, Joulin, Grave, and Lample]{touvron2023llama}
Hugo Touvron, Thibaut Lavril, Gautier Izacard, Xavier Martinet, Marie-Anne Lachaux, Timothée Lacroix, Baptiste Rozière, Naman Goyal, Eric Hambro, Faisal Azhar, Aurelien Rodriguez, Armand Joulin, Edouard Grave, and Guillaume Lample.
\newblock Llama: Open and efficient foundation language models, 2023.

\bibitem[Turpin et~al.(2023)Turpin, Michael, Perez, and Bowman]{turpin2023llmsdontsaywhattheythink}
Miles Turpin, Julian Michael, Ethan Perez, and Samuel Bowman.
\newblock Language models don't always say what they think: Unfaithful explanations in chain-of-thought prompting.
\newblock \emph{Advances in Neural Information Processing Systems}, 36:\penalty0 74952--74965, 2023.

\bibitem[Vaswani et~al.(2017)Vaswani, Shazeer, Parmar, Uszkoreit, Jones, Gomez, Kaiser, and Polosukhin]{vaswani2017}
Ashish Vaswani, Noam Shazeer, Niki Parmar, Jakob Uszkoreit, Llion Jones, Aidan~N. Gomez, Lukasz Kaiser, and Illia Polosukhin.
\newblock Attention is all you need.
\newblock \emph{CoRR}, abs/1706.03762, 2017.
\newblock URL \url{http://arxiv.org/abs/1706.03762}.

\bibitem[Vig et~al.(2020)Vig, Gehrmann, Belinkov, Qian, Nevo, Sakenis, Huang, Singer, and Shieber]{vig2020causal}
Jesse Vig, Sebastian Gehrmann, Yonatan Belinkov, Sharon Qian, Daniel Nevo, Simas Sakenis, Jason Huang, Yaron Singer, and Stuart Shieber.
\newblock Causal mediation analysis for interpreting neural nlp: The case of gender bias.
\newblock \emph{arXiv preprint arXiv:2004.12265}, 2020.

\bibitem[Virtanen et~al.(2020)Virtanen, Gommers, Oliphant, Haberland, Reddy, Cournapeau, Burovski, Peterson, Weckesser, Bright, {van der Walt}, Brett, Wilson, Millman, Mayorov, Nelson, Jones, Kern, Larson, Carey, Polat, Feng, Moore, {VanderPlas}, Laxalde, Perktold, Cimrman, Henriksen, Quintero, Harris, Archibald, Ribeiro, Pedregosa, {van Mulbregt}, and {SciPy 1.0 Contributors}]{2020SciPy}
Pauli Virtanen, Ralf Gommers, Travis~E. Oliphant, Matt Haberland, Tyler Reddy, David Cournapeau, Evgeni Burovski, Pearu Peterson, Warren Weckesser, Jonathan Bright, St{\'e}fan~J. {van der Walt}, Matthew Brett, Joshua Wilson, K.~Jarrod Millman, Nikolay Mayorov, Andrew R.~J. Nelson, Eric Jones, Robert Kern, Eric Larson, C~J Carey, {\.I}lhan Polat, Yu~Feng, Eric~W. Moore, Jake {VanderPlas}, Denis Laxalde, Josef Perktold, Robert Cimrman, Ian Henriksen, E.~A. Quintero, Charles~R. Harris, Anne~M. Archibald, Ant{\^o}nio~H. Ribeiro, Fabian Pedregosa, Paul {van Mulbregt}, and {SciPy 1.0 Contributors}.
\newblock {{SciPy} 1.0: Fundamental Algorithms for Scientific Computing in Python}.
\newblock \emph{Nature Methods}, 17:\penalty0 261--272, 2020.
\newblock \doi{10.1038/s41592-019-0686-2}.

\bibitem[Wang et~al.(2024)Wang, Ge, Shu, Tang, Zhou, He, and Qiu]{wang2024universality}
Junxuan Wang, Xuyang Ge, Wentao Shu, Qiong Tang, Yunhua Zhou, Zhengfu He, and Xipeng Qiu.
\newblock Towards {Universality}: {Studying} {Mechanistic} {Similarity} {Across} {Language} {Model} {Architectures}, October 2024.
\newblock URL \url{http://arxiv.org/abs/2410.06672}.
\newblock arXiv:2410.06672 [cs].

\bibitem[Wang et~al.(2022)Wang, Variengien, Conmy, Shlegeris, and Steinhardt]{wang2022patching}
Kevin Wang, Alexandre Variengien, Arthur Conmy, Buck Shlegeris, and Jacob Steinhardt.
\newblock Interpretability in the wild: a circuit for indirect object identification in gpt-2 small, 2022.
\newblock URL \url{https://arxiv.org/abs/2211.00593}.

\bibitem[Williams(2024)]{williams2024rsacka}
Alex~H. Williams.
\newblock Equivalence between representational similarity analysis, centered kernel alignment, and canonical correlations analysis.
\newblock \emph{bioRxiv}, 2024.
\newblock \doi{10.1101/2024.10.23.619871}.
\newblock URL \url{https://www.biorxiv.org/content/early/2024/10/24/2024.10.23.619871}.

\bibitem[Williams et~al.(2022)Williams, Kunz, Kornblith, and Linderman]{williams2022procrustes}
Alex~H. Williams, Erin Kunz, Simon Kornblith, and Scott~W. Linderman.
\newblock Generalized shape metrics on neural representations, 2022.
\newblock URL \url{https://arxiv.org/abs/2110.14739}.

\bibitem[Wolf et~al.(2019)Wolf, Debut, Sanh, Chaumond, Delangue, Moi, Cistac, Rault, Louf, Funtowicz, et~al.]{wolf2019huggingface}
Thomas Wolf, Lysandre Debut, Victor Sanh, Julien Chaumond, Clement Delangue, Anthony Moi, Pierric Cistac, Tim Rault, R{\'e}mi Louf, Morgan Funtowicz, et~al.
\newblock Huggingface's transformers: State-of-the-art natural language processing.
\newblock \emph{arXiv preprint arXiv:1910.03771}, 2019.

\bibitem[Wu et~al.(2024)Wu, Geiger, Icard, Potts, and Goodman]{wu2024alpacadas}
Zhengxuan Wu, Atticus Geiger, Thomas Icard, Christopher Potts, and Noah~D. Goodman.
\newblock Interpretability at scale: Identifying causal mechanisms in alpaca, 2024.
\newblock URL \url{https://arxiv.org/abs/2305.08809}.

\bibitem[Yamins \& DiCarlo(2016)Yamins and DiCarlo]{Yamins2016}
Daniel~L.K. Yamins and James~J. DiCarlo.
\newblock {Using goal-driven deep learning models to understand sensory cortex}.
\newblock \emph{Nature Neuroscience}, 19\penalty0 (3):\penalty0 356--365, 2016.
\newblock ISSN 15461726.
\newblock \doi{10.1038/nn.4244}.

\bibitem[Yang et~al.(2025)Yang, Li, Yang, Zhang, Hui, Zheng, Yu, Gao, Huang, Lv, et~al.]{yang2025qwen3}
An~Yang, Anfeng Li, Baosong Yang, Beichen Zhang, Binyuan Hui, Bo~Zheng, Bowen Yu, Chang Gao, Chengen Huang, Chenxu Lv, et~al.
\newblock Qwen3 technical report.
\newblock \emph{arXiv preprint arXiv:2505.09388}, 2025.

\bibitem[Zar(2005)]{zar2005spearman}
Jerrold~H Zar.
\newblock Spearman rank correlation.
\newblock \emph{Encyclopedia of Biostatistics}, 7, 2005.

\bibitem[Zhang et~al.(2024)Zhang, Yang, and Agrawal]{zhang2024assessinglearningalignmentunimodal}
Le~Zhang, Qian Yang, and Aishwarya Agrawal.
\newblock Assessing and learning alignment of unimodal vision and language models, 2024.
\newblock URL \url{https://arxiv.org/abs/2412.04616}.

\end{thebibliography}
\bibliographystyle{tmlr}


\pagebreak
\appendix
\section{Appendix / supplemental material}
\begin{figure}[h!]
    \centering
    \includegraphics[width=0.5\textwidth]{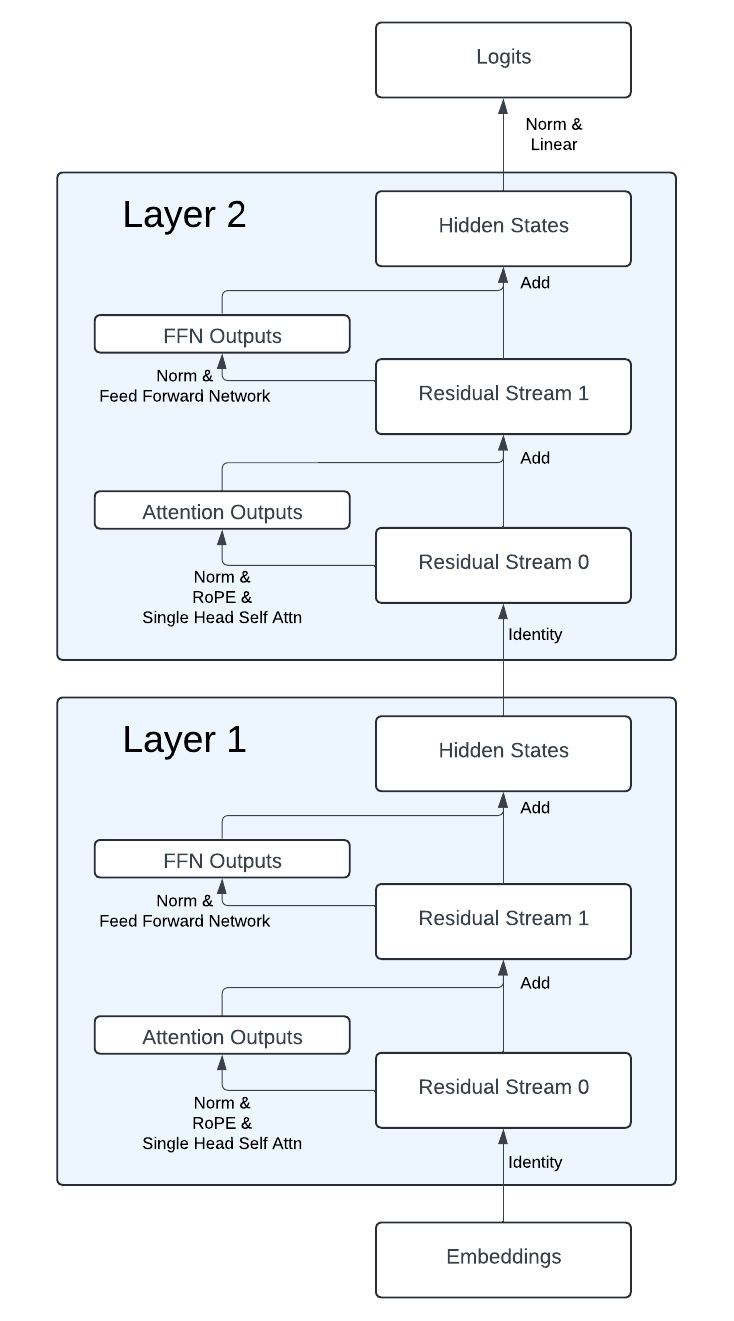}
    \caption{ Figure and caption taken from \cite{grant2024das}.
    Diagram of the transformer architecture used in this work. White
    rectangles represent activation vectors, arrows represent
    functional operations. All causal interventions were performed on either
    the Hidden State activations from Layer 1 or the Embeddings layer. All
    normalizations are Layer Norms \citep{ba2016layernorm}.
    }
    \label{fig:tformerarch}
\end{figure}
\pagebreak

\section{Appendix}

\subsection{Toxicity Experimental Details}\label{sup:toxicity}
\newcommand{\maxlen}{128}
\newcommand{\promptlen}{32}
\newcommand{\genlen}{64}
\newcommand{\toxiclr}{5\times10^{-4}}
\newcommand{\finetunebatchsize}{128}
\newcommand{\numsteps}{10000}
\newcommand{\numgensamples}{1000}
\newcommand{\nummassamples}{10000}

We include an experiment in which we first fine tune a set of DeepSeek-R1-Distill-Qwen-1.5B models \citep{guo2025deepseek} on toxic or nontoxic text completions. We then perform MAS on each model pair to determine whether toxic and nontoxic models are more or less similar to one another.

\subsubsection{Fine-tuning Language Models}

We fine-tuned publicly available DeepSeek-R1-Distill-Qwen-1.5B language models (LMs) \citep{guo2025deepseek} on a concatenation of the train splits of three toxicity-related datasets \citep{jigsaw2018toxiccomments,lin2023toxicchat,bai2022anthropictoxic}. All datasets were filtered to ensure a balanced number of toxic-nontoxic samples before training the model on a specified class (toxic or nontoxic). Samples were considered toxic if any of the possible toxicity levels were true, jailbreaking was true, or the sample was the less favored of two Reinforcement Learning with Human Feedback samples. The filtering consisted of uniformly sampling without replacement from the larger of the two classes until the number of toxic and nontoxic samples reached equality. Training examples were constrained to a maximum input length of $\maxlen$ tokens. Models were optimized using the HuggingFace Trainer API \citep{wolf2019huggingface} with full rank, a learning rate of $\toxiclr$, batch size $\finetunebatchsize$, and a training schedule comprising $\numsteps$ gradient steps. All other training parameters were left to their default values.

\subsubsection{Source Data Collection}

To prepare data for the MAS analyses, we generated seed text samples of length $\promptlen$ from the test splits of the same datasets used for finetuning, similarly after first balancing the number of toxic vs nontoxic samples. We kept both toxic and nontoxic samples for the seed data regardless of the training mode of the model. We then generated $\genlen$ new tokens for each data sample for each model in evaluation mode. To construct the source dataset, we recorded the predicted logits in addition to the model hidden states for each input token during the text generation. The text generation was performed using the argmax over logits for each new token. We collected $\numgensamples$ such prompt-completion samples.

\subsubsection{Toxic LM MAS training details}

To adapt MAS to the aforementioned LM source data, for a given pair of models, we constructed a total of $\nummassamples$ intervention samples from the generated source data where each sample was created by first uniformly sampling a pair of generation samples from each source dataset (with replacement) and then uniformly sampling an intervention index at least 10 tokens from the beginning and ending indices in the sequences (intervention indices ranged between 10 and 54 for the sequences of length 64). We only consider cases using Stepwise MAS, meaning that the same intervention index was used for both the target and source sequences; the source sequence input and output ids and replaced the target input and output ids after the intervention index. To perform an interchange intervention on these samples, we first run the models up to the intervention index, we then perform the intervention from Equation~\ref{eq:interchange}, and then allow the model to continue making predictions from that point onward. We keep the models in evaluation mode for the MAS trainings and we train the rotation matrix by predicting the tokens originally generated by the source model.

\subsection{Model Details}\label{sup:models}
All artificial neural network models were implemented and trained using
PyTorch \citep{pytorch2019} on a single Nvidia Titan X GPU. Unless otherwise
stated, all models used an embedding and hidden state size of
\dmodel\space dimensions. To make the token predictions, each model
used a two layer multi-layer perceptron (MLP) with GELU nonlinearities,
with a hidden layer size of 4 times the hidden state dimensionality with
50\% dropout on the hidden layer. The GRU and LSTM model variants each
consisted of a single recurrent cell followed by the output MLP. Unless
otherwise stated, the transformer architecture consisted of two layers
using Rotary positional encodings  \citep{su2023roformer}. Each model
variant used the same learning rate scheduler, which consisted of the
original transformer \citep{vaswani2017} scheduling of warmup followed
by decay. We used 100 warmup steps, a maximum learning rate of
\learnrate\space, a minimum of 1e-7, and a decay rate of 0.5. We used a
batch size of \batchsize, which caused each epoch to consist of 8
gradient update steps.

\subsection{MAS (and Associated Variants) Training Details}\label{sup:training}
All MAS trainings were implemented and trained using
PyTorch \citep{pytorch2019} on single Nvidia Titan X GPUs.
For each rotation matrix training, we use \dasntrain\space intervention
samples and \dasnval\space samples for validation and testing. We uniformly
sampled corresponding indices upon which to perform interventions, excluding
the B, T, and E tokens in the numeric equivalence tasks from possible
intervention sample indices. In the \arithmetic\space task, we used the comma
token for \varremops\space and \varcumuval\space interventions. When intervening
upon a state in the demo phase in the numeric equivalence tasks, we uniformly
sample a number of steps to continue the demo phase that will keep the object
quantity below \maxcount. We orthongonalize the matrices $Q_{\psi_i}$,
using PyTorch's orthogonal
parametrization with default settings. PyTorch creates the orthogonal matrix
as the exponential of a skew symmetric matrix. We train the rotation matrices
for \dasnepochs\space epochs, with a
batch size of \dasbatchsize\space used for each model index pairing. We only
perform experiments considering two models. Each
gradient step uses the average gradient over batches of all 4 $(i,j)$
ordered pairings.
We select the checkpoint with the best validation performance for analysis.
We use a learning rate of \daslr\space and an Adam optimizer.

For the LSTM architecture, we perform MAS on a concatenation of the $h$
and $c$ recurrent state vectors \citep{hochreiter1997lstm}. In the
GRUs, we operate on the recurrent hidden state. In the transformers, we
operate on the residual stream following the first transformer layer
(referred to as the Layer 1 Hidden States in Supplementary
Figure~\ref{fig:tformerarch}).

Stepwise MAS applies Equation~\ref{eq:interchange} to multiple, contiguous
time-steps in the target and source sequences from step 0 to $u$.
This allows for meaningful transfer of
information in cases of anti-Markovian solutions \citep{grant2024das}.
There is a question of what tokens to use to create each
$h^{(trg)}_{\psi_i}$ before transfer. We show results where each target
token comes from the original target sequence padded by the R token when
$u$ exceeds the target sequence length. See Figure~\ref{fig:masintervention}
for a visualization.


\subsection{RSA Details}\label{sup:rsa}
We performed RSA on a subsample of a dataset of 15 sampled sequences
for each object
quantity ranging from 1-20 on the task that each model was trained on
for each model. We first ran the models on their respective datasets to
collect the latent representations. We sampled 1000 of these latent
vectors as the sample representations in a matrix $M_k \in R^{N\times d_k}$
where $k$ refers to the model index, $N$ is the number of latent vectors
($N = 1000$ in our analyses),
$d_k$ is the dimensionality of a single latent vector for model $k$. We then
calculated the sample cosine distance matrices (1-cosine similarity) for each model resulting in matrices
$C_k \in R^{N\times N}$. Lastly we calculated the Spearman's Rank Correlation
Coefficient between the lower triangles of the matrices $C_1$ and $C_2$
as the RSA value using python's SciPy package
\citep{zar2005spearman,kriegeskorte2008rsa,2020SciPy}. We resampled the 1000 vectors
10 times and recalculated the RSA score 10 times and report the average
over these scores.

\subsection{Representational Similarity Analysis (RSA)}
For a given model layer, we run the model on a batch of sequences consisting of 15
sequences from each object quantity 1-20. We then
sample 1000 representational vectors uniformly from all time points excluding
padding and end of sequence tokens. We construct a Representational Dissimilarity
Matrix (RDM) as 1 minus the cosine similarity matrix over each pair-wise comparison
of the representations (resulting in an RDM of dimensions $1000\times 1000$). We
create an RDM for two models and compare the RDMs using Spearman's rank correlation
on the lower triangle of each matrix \citep{2020SciPy}. We perform the RDM sampling
10 times and report the average over all 10 correlations. See Appendix~\ref{sup:cka}
for CKA methods.

\subsection{CKA Details}\label{sup:cka}
We performed CKA on a subsample of a dataset of 15 sampled sequences
for each object
quantity ranging from 1-20 on the task that each model was trained on
for each model. We first ran the models on their respective datasets to
collect the latent representations. We sampled 1000 of these latent
vectors as the sample representations in a matrix $M_k \in R^{N\times d_k}$
where $k$ refers to the model index, $N$ is the number of latent vectors
($N = 1000$ in our analyses),
$d_k$ is the dimensionality of a single latent vector for model $k$. We then
normalized the vectors along the sample dimension by subtracting the
mean and dividing by the standard deviation (using $d_k$ means and $d_k$
standard deviations calculated over 1000 samples). Using these samples
we calculated the kernel matrices using cosine similarity to create
matrices $C_k \in R^{N\times N}$. Using these matrices, we computed
the Hilbert-Schmidt Independence Criterion (HSIC) where
$I \in R^{N\times N}$ is the identity and $J \in R^{N\times N}$ is a
matrix of values all equal to 1:
\begin{equation}
    H = I - \frac{1}{N} J
\end{equation}
\begin{equation}
    \text{HSIC}(C_1,C_2) = \frac{\text{trace}(C_1 H C_2 H)}{(N-1)^2}
\end{equation}
and lastly we computed CKA as the following:
\begin{equation}
    \text{CKA} = \frac{\text{HSIC}(C_1,C_2)}
        {\sqrt{\text{HSIC}(C_1,C_1) \text{HSIC}(C_2,C_2)}}
\end{equation}
\citep{kornblith2019cka}. We resampled the 1000 vectors
10 times and recalculated the CKA score 10 times and report the average
over these scores.


\begin{figure*}[t!]
    \centering
    \includegraphics[width=0.9\textwidth]{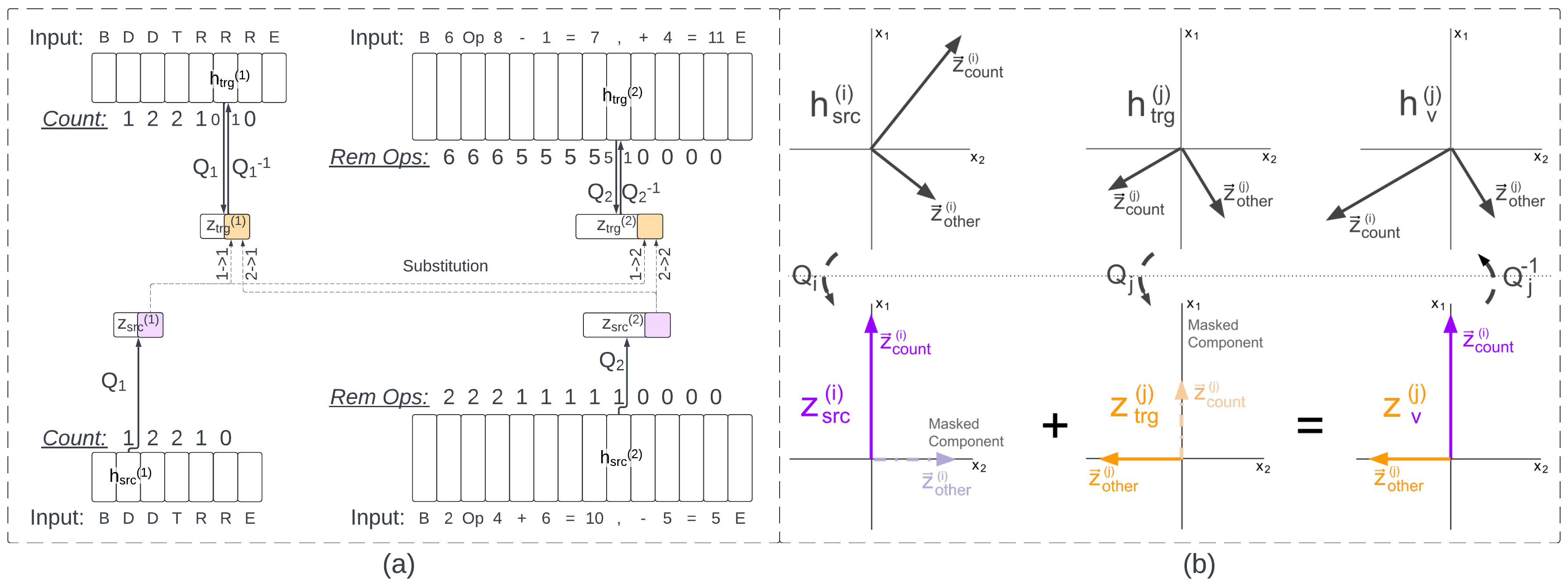}
    \caption{ (a) Diagram of MAS between models trained on
    structurally different tasks. We see all four intervention
    directions on the latent vectors (rectangles) of a
    \multiobject\space GRU, $\psi_1$, and an \arithmetic\space GRU
    $\psi_2$. The value of the causal variables following each
    input token are shown above the $h_{src}$ and below the
    $h_{trg}$ vectors. In the $h_{trg}$ vectors, we see the variable
    value before and after the
    intervention to the left and right of the arrows respectively---the
    \varcount\space for $\psi_1$ and \varremops\space
    for $\psi_2$. The dotted \emph{Substitution} arrows each
    correspond to a single intervention. The models make predictions
    using the intervened vector following the intervention.
    (b) A 2D vector depiction of a hypothetical intervention that
    substitutes the value of the \varcount\space variable from $\psi_i$,
    $\Vec{z}^{\,(1)}_{count}$, into the \varcount\space
    variable, $\Vec{z}^{\,(2)}_{count}$ in $\psi_j$ where the
    superscripts refer to the originating model. Using learned
    matrices $Q_1$ and $Q_2$, $h_{src}^{(1)}$ and $h_{trg}^{(2)}$
    are rotated into $z_{src}^{(1)}$ and $z_{trg}^{(2)}$ where the
    $\Vec{z}_{count}$ subspace is organized into a contiguous
    subset of vector dimensions disentangled from all other
    information. In this aligned space the $\Vec{z}_{count}$ values
     can be freely exchanged without affecting
    other information. Lastly, $z_{\text{v}}^{(2)}$
    is returned to $\psi_2$'s hidden state space by inverting
    $Q_2$ and is used for inference by $\psi_2$.
    }
    \label{fig:masdiagram}
\end{figure*}
\subsection{Arithmetic}\label{sup:arithmetic}
To further explore how representations of number differ across tasks,
we include a comparison between GRUs trained on the \multiobject\space
task and an arithmetic task. See Figure~\ref{fig:masdiagram} for
a visual of MAS applied between the \arithmetic\space and \multiobject\space
tasks.

\textbf{Arithmetic Task:}
This task consists of an indication of the number of addition/subtraction
operations, an initial value, and then the operations interlaced
with the cumulative value (cumu val) of the operations. An example sequence is:
"B 3 4 + 3 = 7 , + 11 = 18 , - 5 = 13 E", where the first number
indicates the number of operations in the trial. The 2nd number is a
sampled start value. Each operand is a sampled value that is combined with
the cumu val according to the operator. The number of
operations is uniformly sampled from 1-10. The starting value is uniformly
sampled from 0-20. All numeric values are restricted inclusively to 0-20.
The operations are uniformly sampled from $\{+,-\}$ when the
cumu val is in the range 1-19. When 0 and 20 the operations are
"+" and "-" respectively. The operands are uniformly sampled from the set
that restricts the next cumu val to the range 0-20.
The cumu val is shown after the "=" for each operation. The
"," token follows the cumu val until the end of the sequence in which the
E token replaces the comma. The model must predict the "=" tokens,
cumu vals, commas, and the E token for a trial to be correct. We use a base
21 token system so that all values correspond to a single token.

\begin{figure}[t!]
    \centering
    \includegraphics[width=.5\columnwidth]{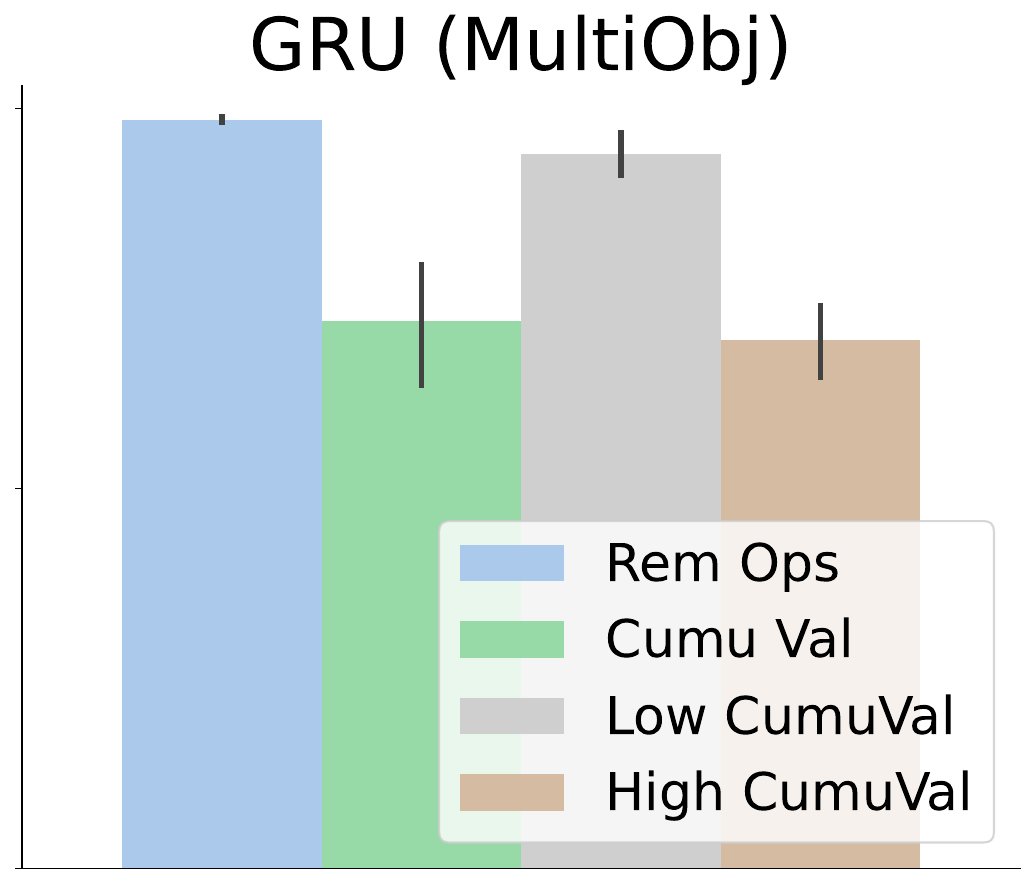}
    \caption{
    MAS used to compare the \varcount\space variable from
    the \multiobject\space GRUs to the \varremops\space and
    \varcumuval\space variables from the \arithmetic\space GRUs.
    \varcumuval\space results are from MAS trained with all possible
    \varcumuval\space values. Low \varcumuval\space results are from a
    separate MAS analysis restricted to \varcumuval\space values of
    1-10, same as the range of \varremops\space variable.
    High \varcumuval\space results are from MAS conditioned on values
    from 11-20.
    }
    \label{fig:arithresults}
\end{figure}

\textbf{Arithmetic Results}
We include a MAS analysis between \arithmetic\space GRUs and GRUs trained on
the Numeric Equivalence tasks (see Figure~\ref{fig:arithresults}).
The leftmost panel shows that we can successfully align the \varcumuval,
and the \varremops\space variables between and within the \arithmetic\space
GRUs. The middle panel shows that MAS can successfully align the \varcount\space
with the \varremops\space variables between GRUs trained on the
\multiobject\space and \arithmetic\space tasks respectively. These results
are qualified by the lower IIA alignment between the \varcount\space and the
\varcumuval\space variables. We see that when we perform MAS only
on \varcumuval\space values that are shared with possible the \varremops\space
values (Low CumuVal), the results are much higher but still do not match the
results from \varremops. These findings are consistent with the hypothesis that
these GRUs are using different types of numeric representations for arithmetic
than incremental counting.

\subsection{Symbolic Variables}\label{sup:variables}
Inspired by DAS, MAS has
the ability to find alignments for specific types of information by conditioning
the counterfactual sequences specific causal variables (i.e. the count of the
sequence in the numeric equivalence tasks).

In all numeric equivalence tasks, we prevent interventions on representations
resulting from the BOS, T, and EOS tokens. In the arithmetic task, we only
perform interventions on representations after the "," token.
We perform MAS using each of the following causal variables,
where the corresponding task is denoted in parentheses:
\begin{enumerate}
    \item \textbf{Full} (Arithmetic/Num Equivalence/\modulo/\rounding): Refers to cases in which we
    transfer all causally relevant information between models (not all activations).
    \item \textbf{Count} (Num Equivalence/\modulo/\rounding): The
    difference between the number of observed demo tokens and the number of
    response tokens in the sequence. Example: the following sequences have a
    Count of 2 at the last token: "B D D" ; "B D D D T R"
    \item \textbf{Count} (\modulo): The number of observed demo tokens when
    in the demo phase, and modulo \modulus\space of the number of demo tokens
    minus the number of response tokens when in the response phase. Example:
    the following
    sequences have a Count of 2 at the last token: "B D D" ; "B D D D D D D D T R"
    \item \textbf{Count} (\rounding): The number of observed demo tokens when
    in the demo phase, and the number of demo tokens rounded to the nearest
    multiple of \roundmultiple\space minus the number of response tokens when
    in the response phase. Example: the following
    sequences have a Count of 2 at the last token: "B D D" ; "B D D D D T R"
    \item \textbf{\varcumuval} (Arithmetic): The cumulative value of the
    arithmetic sequence in the Arithmetic task. Example: if we substitute in
    a value of 3 at the "," token in the sequence "B 2 Op 3 + 5 = 8 ,"
    the counterfactual sequence could be "B 2 Op 3 + 5 = 8 , + 2 = 5 E".
    \item \textbf{\varremops} (Arithmetic): The remaining number of operations
    in the arithmetic sequence. Example: we substitute in a value of 1 at the
    "," token in the sequence "B 3 Op 3 + 5 = 8 ,", the counterfactual sequence
    could be "B 3 Op 3 + 5 = 8 , + 1 = 9 E".
\end{enumerate}


\end{document}